\documentclass[sigconf, nonacm]{acmart}

\renewcommand\footnotetextcopyrightpermission[1]{}
\acmConference[]{}{}{}  %

\AtBeginDocument{%
  \providecommand\BibTeX{{%
    \normalfont B\kern-0.5em{\scshape i\kern-0.25em b}\kern-0.8em\TeX}}}

\usepackage{xspace}

\newcommand{\figref}[1]{Fig.~\ref{#1}} 
\newcommand{\ie}{\textit{i.e.}\xspace} 
 \newcommand{\etal}{\textit{et
al.}\xspace}
\newcommand{\wrt}{w.r.t\xspace}

\newenvironment{tightitemize}{%
\begin{list}{$\bullet$}{%
\setlength{\itemsep}{1.5pt}%
\setlength{\topsep}{2pt}%
\setlength{\parskip}{0pt}%
\setlength{\parsep}{0pt}%
 \setlength{\labelwidth}{0pt}%
\setlength{\leftmargin}{4pt}%
\setlength{\labelsep}{0pt}%
\setlength{\listparindent}{0pt}%
}}%
{\end{list}}

{\end{list}}

\usepackage{xcolor}
\usepackage{xspace}
\usepackage{tabularx}

\definecolor{codegreen}{rgb}{0,0.6,0}
\definecolor{codegray}{rgb}{0.5,0.5,0.5}
\definecolor{codepurple}{rgb}{0.58,0,0.82}
\definecolor{backcolour}{rgb}{0.95,0.95,0.92}
\definecolor{purple}{RGB}{128,0,128}
\definecolor{indigo}{RGB}{75,0,130}
\definecolor{royalblue}{RGB}{65,105,225}
\definecolor{navy}{RGB}{0,0,128}
\definecolor{codebrown}{rgb}{0.6,0.6,0}

\newif\ifcommenton 
\commentonfalse %
\ifcommenton

\newcommand{\todo}[1]{\textcolor{red}{TODO: #1}}
\newcommand{\ada}[1]{\textcolor{orange}{AG: #1}}
\newcommand{\alexey}[1]{\textcolor{indigo}{AT: #1}}
\newcommand{\mlee}[1]{\textcolor{magenta}{ML: #1}}
\newcommand{\dhruv}[1]{\textcolor{purple}{DG: #1}}
\newcommand{\ds}[1]{\textcolor{codegreen}{DS: #1}}
\newcommand{\neha}[1]{\textcolor{royalblue}{NL: #1}}
\newcommand{\vis}[1]{\textcolor{royalblue}{VR: #1}}

\else
\newcommand{\todo}[1]{}
\newcommand{\ada}[1]{}
\newcommand{\alexey}[1]{}
\newcommand{\mlee}[1]{}
\newcommand{\dhruv}[1]{}
\newcommand{\ds}[1]{}
\newcommand{\neha}[1]{}
\newcommand{\vis}[1]{}
\fi

\usepackage[utf8]{inputenc}
\usepackage{subcaption}
\usepackage{caption}
\usepackage{pifont}
\usepackage{graphicx}
\usepackage{array}
\usepackage{multirow}
\usepackage{lipsum}
\usepackage{float}
\usepackage{makecell}
\usepackage{algorithm}
\usepackage{amsmath}
\usepackage{algpseudocode}
\usepackage[normalem]{ulem}
\newcommand{\sys}{FeLiX\xspace}
\newcommand{\sysAware}{TierTrack\xspace}
\newcommand{\sysSample}{TierSelect\xspace}
\newcommand{\sysAggregate}{TierFuse\xspace}

\newcommand{\cpEval}{\textit{AVL\_EVAL}\xspace}
\newcommand{\cpTrain}{\textit{AVL\_TRAIN}\xspace}
\newcommand{\cpUnavl}{\textit{UN\_AVL}\xspace}

\newcommand{\Oort}{OORT\xspace}
\newcommand{\OortStar}{\textit{OORT$\ast$}\xspace}
\newcommand{\OortAsync}{\textit{OORT+Async}\xspace}
\newcommand{\OortAsyncStar}{\textit{OORT+Async$\ast$}\xspace}
\newcommand{\ReflStar}{{REFL$\ast$}\xspace}

\usepackage{tikz}
\newcommand*\circled[1]{\tikz[baseline=(char.base)]{
            \node[shape=circle,draw,inner sep=1pt] (char) {#1};}}

\newcommand{\myparagraph}[1]{\noindent{\bfseries #1.}}

\begin{document}

\title[Robust Federated Learning Under Real-World Client Churn]{Robust Federated Learning Under Real-World Client Churn}

\author{Dhruv Garg}
\email{dgarg39@gatech.edu}
\affiliation{\institution{Georgia Tech}\country{USA}}

\author{Neha Lakhani}
\email{nlakhani9@gatech.edu}
\affiliation{\institution{Georgia Tech}\country{USA}}

\author{Debopam Sanyal}
\email{dsanyal7@gatech.edu}
\affiliation{\institution{Georgia Tech}\country{USA}}

\author{Myungjin Lee}
\email{myungjle@cisco.com}
\affiliation{\institution{Cisco Research}\country{USA}}

\author{Alexey Tumanov}
\email{atumanov@gatech.edu}
\affiliation{\institution{Georgia Tech}\country{USA}}

\author{Ada Gavrilovska}
\email{ada@cc.gatech.edu}
\affiliation{\institution{Georgia Tech}\country{USA}}

\renewcommand{\shortauthors}{Garg, Lakhani, Sanyal, Lee, Tumanov, Gavrilovska}

\begin{abstract}
Federated Learning (FL) enables training shared models on private, on-device
data, yet current production systems are typically constrained to slow,
multi-day refresh cycles due to the complexity of coordinating massive client
populations. However, for modern applications such as feed ranking,
ad-targeting, and personalized recommendation systems, model \emph{freshness},
the ability to adapt rapidly to new user-local data, is a critical driver for
maximizing application objectives like click-through-rate. This temporal gap
leaves models stale and unresponsive to volatile live data distributions, such
as viral trends or shifting user intents. Bridging this gap requires navigating
three interlocking constraints that existing FL systems don't address: transient
client availability, highly dynamic data heterogeneity, and the inherent delay
between model predictions and observable application outcomes. We present
\textbf{\sys}, a redesign of FL orchestration that optimizes
\emph{time-to-target} accuracy on live interaction streams. \sys introduces
three novel primitives: (i) \textit{streaming-aware availability tiers}, which
use lightweight telemetry to surface ready clients at scale; (ii)
\textit{fresh-utility selection}, a dual-tier mechanism that scouts for
high-value statistical contributions on devices capable of meeting tight refresh
deadlines; and (iii) \textit{informativeness-aware, delay-robust aggregation},
which incorporates late, high-value updates containing ground-truth outcomes
without biasing the global model toward obsolete distributions. Unlike existing
systems that rely on unrealistic oracular knowledge of client availability, \sys
achieves near-oracular performance in real-world settings. Evaluated across
multiple modalities (CIFAR-10 and Google Speech) and realistic low-availability
traces, \sys reduces wall-clock time-to-target accuracy by up to $2.37\times$
while also saving $1.30\times$ total communication bandwidth compared to
state-of-the-art Synchronous and Asynchronous FL baselines. By enabling faster
model adaptation under adverse real-world conditions, \sys ensures streaming
applications remain in sync with the most recent user interactions.
\end{abstract}

\maketitle

\section{Introduction}
\label{sec:intro}

Federated Learning (FL) enables training shared models across large populations
of user devices without collecting the raw data centrally. This paradigm has
unlocked machine learning on private, on-device interaction logs that were
previously inaccessible to centralized trainers
\cite{mcmahan2017communication,bonawitz2019flsysdesign,kairouz2021advances}. Such privacy-preserving
decentralization is now critical for various production applications, including
personalized recommenders, feed-ranking, and session-aware input prediction.
FL has been deployed for mobile keyboard and language prediction~\cite{hard2018federated,chen2019federated,xu2023federated},
privacy-preserving healthcare analytics~\cite{brisimi2018federated,xu2021federated,nguyen2022federated},
and large-scale model personalization~\cite{yang2019federated,xu2024fwdllm},
establishing it as the de facto approach for privacy-sensitive collaborative training across diverse domains.
Historically, these deployments focused on batched, multi-day refreshes where
modest delays between data collection and model updates were acceptable.

Contemporary on-device applications, however, increasingly require \emph{fresh}
models that adapt on hourly timescales. User interaction streams are volatile:
viral content, news events, promotional campaigns, and short-lived user intents
can shift predictive distributions rapidly. When model refresh intervals lag
behind these shifts, predictive relevance degrades and key application metrics
such as click-through rates and downstream conversions suffer
\cite{matam2024quickupdate}. Meeting these goals demands FL systems that
minimize \emph{time-to-target accuracy} on live streams while operating at
production scale.

\begin{figure}[t!]
    \centering
    \begin{subfigure}[b]{0.98\columnwidth}
        \centering
        \includegraphics[width=\textwidth]{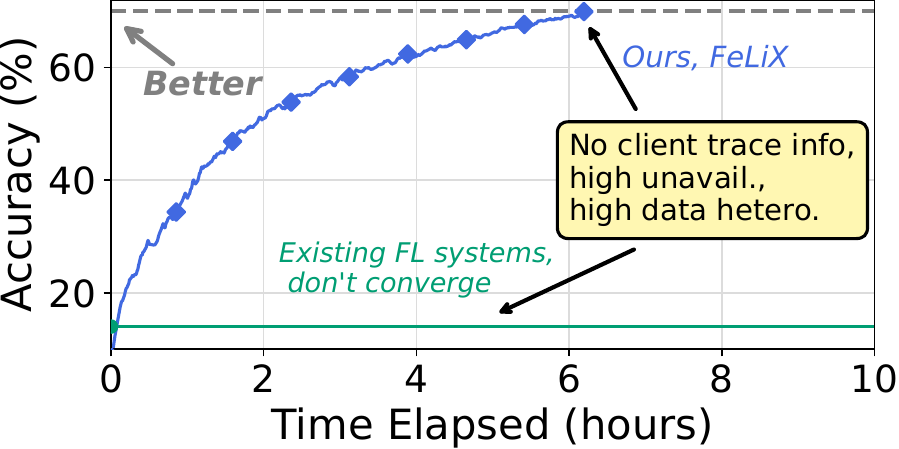}
        \label{fig:intro:felix-existing-gap}
    \end{subfigure}
    \caption{Existing FL systems stall, mis-select, or accumulate staleness under transient availability and heterogeneous streaming data, whereas \sys continuously trains and reaches target accuracy without trace knowledge through adaptive, low-overhead orchestration.}
    \label{fig:intro:felix-existing-gap}
\end{figure}

Transitioning from slow and batched to agile and streaming FL is difficult
because near-real-time updates break the core assumptions of existing FL
systems. We highlight three interlocking constraints that render naive
extensions of current architectures ineffective:

\paragraph{(1) Transient client availability.} Mobile and edge devices are
intermittently reachable due to battery, network, and user behavior.
Availability fluctuates at minute-level granularity and exhibits heavy churn
\cite{mobiperf}. As shown in~\cite{garg2025client}, existing mitigations generally fail in three ways: (a)
\emph{Over-selection}, where requesting excess clients to participate in each
round wastes device compute and bandwidth while increasing server-side update
variance; (b) \emph{blocking synchrony}, which converts per-round latencies from
seconds to minutes when aggregation goals are not met; and (c) \emph{coarse
periodic probing}, which either misses short-lived availability windows of
useful clients or incurs high heartbeat overhead to maintain up-to-date status.
These approaches trade away either time-to-accuracy or communication budgets,
making them unsuitable for large-scale FL deployments operating with thousands
of clients per round.
Approaches that target the statistical variance from partial participation~\cite{jhunjhunwala2022fedvarp}
or study large-cohort selection effects~\cite{charles2021large} improve efficiency
but still assume a predictable pool of available clients and do not address minute-level churn.

\paragraph{(2) Evolving client utility and heterogeneity.} Clients differ not
only in their hardware capabilities (compute speed, memory, network), but also
in the \emph{statistical value} of their recent data. We define \emph{client
utility} as a time-varying quantity combining data informativeness for the live
distribution with the client's expected compute latency. Existing selection
algorithms often ignore these dynamics
\cite{nguyen2022fedbuff,bonawitz2019flsysdesign,nishio2019client}, while newer ranking methods
rely on the "oracular" knowledge of availability, which is impractical in
deployment. Not just that, they don't account for the shifting data
distributions of streaming tasks \cite{lai2021oort,abdelmoniem2023refl,ye2023heterogeneous}.
Selectors that rank clients using utility estimates derived from infrequent
probes or historical traces systematically mis-rank participants, as these
signals quickly become stale and introduce bias that can outweigh their intended
benefits. For example, it may over-select fast devices with irrelevant data
while neglecting slow devices with fresh, informative signals. Such techniques
either require non-existent oracular deployment traces, coarse-grained and often
stale utility estimates, or frequent probing that incurs prohibitive overhead.
As a result, they suffer rapid degradation in time-to-accuracy with shifts in
utility.

\paragraph{(3) Model freshness vs.\ delayed updates.} In production tasks, the
strongest training signal (the ground-truth label) often arrives with some
delay; for instance, a recommendation's conversion may be observed hours after
the interaction. This creates a fundamental trade-off: updating immediately with
proxy feedback maximizes freshness but introduces noise, while waiting for full
feedback improves label quality at the cost of staleness. Existing strategies—like buffered asynchronous aggregation~\cite{nguyen2022fedbuff}
or staleness-weighting schemes~\cite{xie2019asyncFLopt,rodio2024fedstale,zhou2022towards}—force
a choice between convergence speed and statistical robustness. Incorrect aggregation can destabilize the training process,
inflating the wall-clock time required to reach target accuracy or discarding
valuable, though delayed, signals.

Taken together, these constraints explain why existing synchronous and
asynchronous FL systems fail to achieve reasonable time-to-target accuracy in
streaming settings, at production scale as shown in
~\figref{fig:intro:felix-existing-gap}. Synchronous schemes stall or
over-select; asynchronous schemes avoid blocking but accumulate staleness that
degrades convergence~\cite{xie2019asyncFLopt,lian2018asynchronous,li2019convergence};
assumed oracular knowledge of availability or heavy-weight
predictors are not tractable; and prohibitive communication costs of frequent
probes derail performance. Empirical studies (see
\S\ref{sec:background_motivation} and
~\figref{fig:background:hetero_unavail_baselines_gap_plot}) show that even
modest unavailability or data heterogeneity can worsen the model adaptation time
as the complexity of the deployment shifts to hardened production settings. Not
just that, it can substantially degrade performance, even to the point of non
convergence, as the unavailability and heterogeneity increase. 

In this work, we frame streaming FL as a constrained optimization problem:
\emph{minimize time-to-target accuracy on a live interaction stream subject to
transient availability, evolving per-client utility, frequent update delays, and
bounded communication overhead.} This objective focuses our design on policies
that treat availability and utility as low-latency control inputs while ensuring
scalability across massive populations, without assuming oracular knowledge of
client behaviour.

We present \textbf{\sys}, a redesigned FL orchestration framework that directly
optimizes \textit{time-to-target accuracy} in streaming production settings.
\sys treats availability as an evolving signal and couples inexpensive freshness
checks with selection logic to optimize convergence. Our core primitives are:

\begin{tightitemize}
    \item \textbf{Streaming-aware availability tiers:} \sys organizes clients
    into dynamic readiness tiers based on lightweight signals (battery,
    connectivity). This leverages a client-server design that minimizes network
    overhead while ensuring the server possesses high-fidelity information on
    client capabilities within tight refresh windows.
    \item \textbf{Fresh-utility based selection:} We introduce a dual-tier
    approach where an \emph{evaluation tier} performs inexpensive forward passes
    to estimate data informativeness. This is fused with compute speed to
    prioritize clients that are both hardware-performant and reflect the most
    relevant recent distribution shifts.
    \item \textbf{Informativeness-aware, delay-robust aggregation:} \sys employs
    weights that balance update informativeness with model-age corrections. This
    allows the system to safely incorporate late, high-value updates with full
    feedback, ensuring the global model remains robust without being stalled by
    the long-tail latencies of the straggler or long unavailable clients.
\end{tightitemize}

We conduct an exhaustive evaluation of \sys using real-world MobiPerf traces and
multi-modal datasets including CIFAR-10~\cite{Krizhevsky09learningmultiple} and Google Speech Commands V2~\cite{speechcommandsv2}. We
demonstrate that \sys achieves a $2.37\times$ reduction in time-to-target
accuracy compared to recent Synchronous and Asynchronous FL baselines, while
limiting communication overhead and preserving the predictive power of fresh
models.

\section{Background and Motivation}
\label{sec:background_motivation}

\begin{figure}[t]
    \centering
    \begin{subfigure}[b]{0.6\columnwidth}
        \centering
        \includegraphics[width=\textwidth]{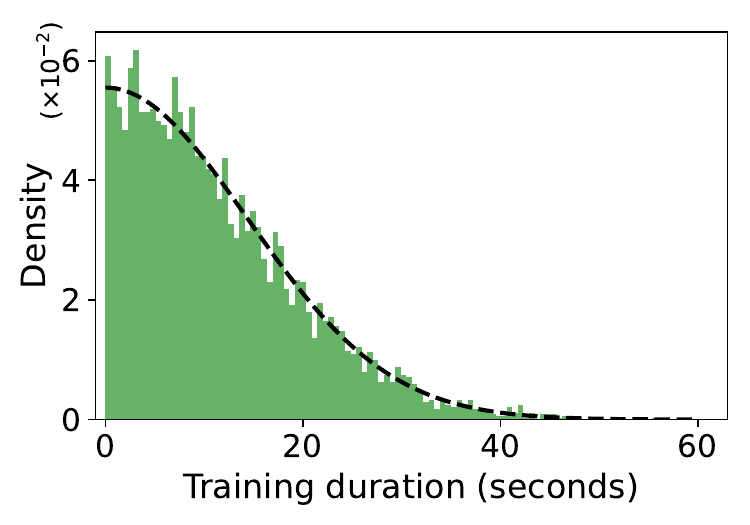}
    \end{subfigure}
    \caption{Cross-device trainers exhibit 2–120 s execution times driven by
        hardware, data, and WAN variability~\cite{nguyen2022fedbuff}. While
        stragglers delay updates through slow execution, transient device
        unavailability pauses participation entirely, highlighting a key systems
        challenge beyond conventional straggler mitigation.}
    \label{fig:motiv:fedbuff_training_delay}
\end{figure}

\begin{figure}[t]
    \centering
    \begin{subfigure}[b]{0.5\columnwidth}
        \centering
        \includegraphics[width=\textwidth]{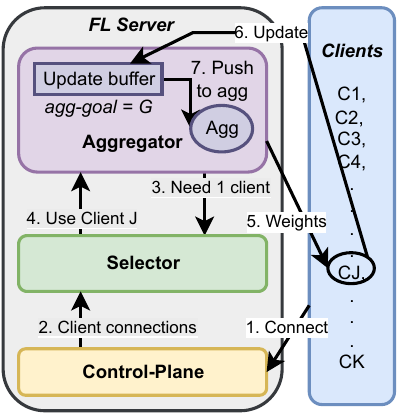}
        \label{fig:background:asyncfl_existing}
    \end{subfigure}
    \caption{Overview of cross-device AsyncFL training architecture. To
        meet the \textit{concurrency level}, clients are asynchronously selected
        at random to train locally and communicate updates to the server. The
        server aggregates the updates to get the new global model whenever
        \textit{agg-goal} number of clients have sent updates.}
    \label{fig:background:asyncfl_existing_system}
\end{figure}

\subsection{FL Primer and Client Availability}
\label{sec:background:primer_runtime}

\myparagraph{Asynchronous FL}
Federated Learning (FL) trains a shared model across many decentralized,
privacy-sensitive clients by keeping raw data local: the server broadcasts
global weights, clients perform local training on-device, and the server
aggregates returned updates
\cite{mcmahan2017communication,bonawitz2019flsysdesign}. Two execution regimes
dominate cross-device deployments. In \emph{synchronous} FL (SyncFL) the server
advances in rounds after collecting an \emph{agg-goal} number of client updates;
in \emph{asynchronous} FL (AsyncFL) the server incorporates updates as they
arrive and does not block on stragglers. SyncFL offers conceptual simplicity and
bounded staleness but is vulnerable to delays caused by a few slow or
unavailable clients; AsyncFL trades that blocking for continuous progress at the
cost of needing principled staleness handling.
Convergence analyses of FedAvg under non-IID data~\cite{li2019convergence,wang2020tackling}
reveal that data heterogeneity exacerbates gradient divergence across clients,
while adaptive federated optimizers~\cite{reddi2020adaptive} partially compensate
but leave the availability challenge unaddressed.

AsyncFL emerged to address the wide runtime variability observed in practice:
theoretical analyses of asynchronous decentralized SGD~\cite{lian2018asynchronous}
and asynchronous federated optimization~\cite{xie2019asyncFLopt} confirm that
async updates converge under mild staleness bounds but can degrade sharply when
updates are very stale or derived from strongly non-IID local data.
mobile devices exhibit long-tailed training runtimes driven by differences in
compute, background load, and network latency as shown
in~\figref{fig:motiv:fedbuff_training_delay}. Together, device heterogeneity and
application-driven data generation produce a rapidly evolving trainer runtime
distribution. This variability forces system designers to confront three
co-occurring concerns: (i) which clients should the server select at any moment,
(ii) how to cheaply detect that a client is actually ready to train, and (iii)
how to incorporate late or intermittent updates without destabilizing learning.
A compact AsyncFL control loop (selector, weight dispatcher, client training,
aggregator/buffer) exposes these decision points; we illustrate the flow in
\figref{fig:background:asyncfl_existing_system}. Clients register themselves
with the FL-server \circled{1}. The control-plane maintains these connections
and forwards the client-id(s) to the selector \circled{2}. The selector is
tasked with maintaining a desired training concurrency (\ie, number of parallely
training clients) throughout the process. Whenever the training count falls
below the concurrency level (\ie, clients return updates). The selector uses its
algorithm to return the next client(s) to train \circled{4} so as to meet the
concurrency level. These selected client ID(s) are forwarded to the aggregator.
\circled{5} The weight dispatcher sends the latest server model weights (W) to
the selected clients. If the client is available to train, it begins training
using fresh global weights (W) and data from the applications' local data-store.
Upon completion, it returns weight changes \wrt W as updates ($\Delta W$)
\circled{6} to the aggregator. If the client wasn't initially available to train
or is interrupted while training, the client waits until it becomes available
next. At the FL server, the model updates received from clients are aggregated
and put into a temporary buffer. Once agg-goal number of updates in the buffer
are aggregated, the server model version changes and the buffer is emptied
\circled{7}. This FL training process continues until convergence.

\myparagraph{Evolving runtime availability} While foundational FL research and
early system prototypes often assumed highly available clients operating in
stable, conducive environments, production deployments reveal that this
assumption is fundamentally flawed. In practice, ignoring the reality of
intermittent participation degrades performance significantly, inflating
time-to-accuracy and wasting system resources. We define \emph{availability} not
merely as a device being powered on, but as a multi-dimensional state meeting
strict system-level and application-level filters, including WiFi-only
connectivity, sufficient battery levels, and OS-mandated idle constraints.
Consequently, \emph{unavailability} is the norm; production traces from the
Google FL System~\cite{bonawitz2019flsysdesign}, the PAPAYA analytics
platform~\cite{srinivas2025papaya}, LinkedIn’s FLINT~\cite{wang2023flint}, and
FedScale/MobiPerf~\cite{lai2022fedscale,mobiperf} report mean client
availability ranging from as low as \texttt{15\%} to roughly \texttt{50\%} of
the total population. In production, \textit{unavailability is therefore the
norm rather than the exception.} This participation is inherently
\emph{transient}, characterized by short-lived ready windows whose durations are
often comparable to the training round itself. Across different geographies and
applications, these dynamics manifest in large peak-to-trough swings. For
instance, diurnal effects can cause availability to fluctuate by $5\times$
within a single day, with minima dropping to \texttt{10\%} during active usage
hours and peaking at \texttt{80\%} when devices are idle and charging. At the
system level, these trends create significant orchestration challenges: in the
Google FL system, nearly \texttt{22\%} of updates arrive too late for
aggregation and are rejected, while another \texttt{2\%} are interrupted
mid-computation. Restrictive participation criteria, while necessary for user
experience, further contract the eligible pool, sometimes as low as
\texttt{22\%} of the total devices as shown in MobiPerf~\cite{mobiperf}. This
exposing a sharp trade-off between model fairness and system throughput. These
bursty and unpredictable availability patterns mean that naive selection
strategies either over-select to mask dropouts (increasing variance) or stall
rounds waiting for stragglers, both of which are untenable for streaming
applications requiring hourly refreshes.
Prior measurements~\cite{garg2025client} directly quantify this: even under
moderate unavailability ($<$50\%), time-to-target accuracy degrades sharply,
confirming that availability must be treated as a first-class orchestration
signal rather than a background metric.

\myparagraph{AsyncFL alone cannot fix unavailability} These runtime and
availability characteristics translate into three practical operational effects
that motivated AsyncFL and that now motivate streaming-aware controls. First,
servers often resort to \emph{over-selection} (invite many more candidates than
needed) to probabilistically meet an agg-goal; this wastes device CPU,
uplink/downlink network bandwidth and increases variance in server-side latency.
Second, a client's local training time may exceed the mean aggregator interval,
increasing the chances of interruptions, rejected updates, or partial work.
A natural mitigation is to checkpoint local training state so that a client can
resume after an interruption: an approach well-studied in distributed DNN
training~\cite{mohan2021checkfreq,eisenman2022checkNRun,gupta2024jitCheckpt,wang2023geminiRecovery}.
However, in the FL context, checkpointing only preserves progress on the
\emph{client}; the client still cannot return any update until device conditions
recover, and the global model advances in the meantime, rendering the saved
gradients stale on return.
Checkpointing therefore does not close the availability gap. It defers the
problem rather than solving it.
Third, restrictive eligibility filters shrink the effective population and risk
selection bias. Together, these effects show that availability and runtime
behavior are not background metrics but \emph{low-latency control inputs} that
selection and aggregation logic must use in order to optimize wall-clock
progress and maintain model quality.

Finally, while AsyncFL and its improvements such as update buffering reduce
blocking, they do not by themselves solve the coupled problems of rapid
availability churn, evolving per-client statistical value, and bounded
communication costs. For example, buffered AsyncFL reduces immediate stalls but
requires careful eviction/weighting policies to prevent stale updates from
hurting convergence. Later sections discuss these algorithmic and systems
tensions in detail and show why treating readiness and freshness as first-class,
lightweight signals is essential for streaming FL.

\subsection{FL Orchestration Challenges}
\label{sec:background:sota}

\myparagraph{Brittle selection algorithms}
Many FL selection strategies attempt to balance synchronization and
participation, but practical deployments expose a harsh trade-off between update
staleness and system stalls. Synchronous FL variants avoid staleness by
enforcing global rounds but are highly vulnerable to \emph{stalls}: when
availability is intermittent, SyncFL blocks on stragglers or turns to heavy
over-selection, converting per-round latencies from seconds into minutes and
dramatically increasing communication and cost. Asynchronous and buffered
schemes relax blocking by accepting updates computed on older model states (or
by aggregating along a buffer), which improves throughput. However, if stale
updates are naively aggregated, they will destabilize convergence. Conversely,
discarding late updates sacrifices the valuable statistical signal carried by
the slower or transiently available clients. Recent selector-based approaches
that prioritize high-utility clients (for example, OORT and REFL selectors) or
use learned predictors to estimate client utility aim to trade off speed and
signal by favoring fast, high-utility workers. In practice, these methods
typically depend on availability signals or historical traces to estimate
utility. Under bursty availability and rapid population/data shifts those
utility estimates age quickly leading to repeated mis-ranking. For example, such
selecots would over-select fast but irrelevant devices and neglect informative
but ephemeral clients. These limitations have motivated hybrid approaches like
buffered asynchronous aggregation that attempt to retain long-tail information
while reducing stalls, but the tension between timeliness, stability, and
privacy-compatible aggregation remains unresolved.
Variance-reduction techniques for partial participation~\cite{jhunjhunwala2022fedvarp}
and large-cohort selection~\cite{charles2021large} target statistical stability
but still assume a stable, predictable client pool.
Staleness-aware schemes~\cite{rodio2024fedstale,zhou2022towards} weight or
discount late updates based on model-version gap, but do not resolve the
challenge of detecting and exploiting short availability windows in real time.

\myparagraph{Coarse systems and instrumentation}
Production stacks provide several primitives for large-scale FL, but their
availability signals tend to be either too coarse to support fine-grained
orchestration or too heavyweight to scale. Such platforms use techniques like
client check-ins, tokens, and over-selection heuristics to mask dropout and
reduce orchestration complexity with client availabilty. Yet such protocols are
explicitly designed around coarse-grained and slower tradeoffs like
accept/reject tokens, windowed selection rather than continuous, low-latency
availability views. This design choice simplifies robustness but also means a
nontrivial fraction of potential updates are delayed or rejected in practice. At
the other extreme, tooling that instruments devices more deeply like
pre-deployment measurement, capability profiling, or continuous heartbeat
reporting, can provide the fine-grained, near-real-time availability signals
orchestration needs, but at production scale these signals increase reporting
overhead, consume device energy, and raise privacy and operational concerns.
Benchmark and integration frameworks make clear how availability and
distribution shift in deployment. They also let researchers replay realistic,
time-varying traces for evaluation. However, by definition these offline traces
cannot supply the compact, cross-application, real-time view an orchestrator
would need for streaming, hourly refresh tasks. In short, production systems
either expose high-fidelity signals that are too heavy to maintain continuously,
or coarse signals that force servers to choose between blocking (stalling
progress) and proceeding blindly (accumulating staleness). Representative system
and benchmarking efforts that highlight these tradeoffs include Google
FLSys~\cite{bonawitz2019flsysdesign} and the PAPAYA analytics
stack~\cite{srinivas2025papaya}, which describe token/check-in and rejection
mechanics for large-scale mobile FL; the benchmark suite
FedScale~\cite{lai2022fedscale}; and the FL integration tooling
FLINT~\cite{wang2023flint}, which emphasizes on-device measurement and
pre-deployment instrumentation.
Research frameworks such as Flower~\cite{beutel2020flower} and
FedML~\cite{he2020fedml} enable flexible experimentation but are designed for
controlled settings rather than production-scale streaming deployments with
bursty, minute-level availability dynamics.

These algorithmic and system-level gaps are particularly acute for
\textit{streaming applications}, where the requirement for hourly refreshes
clashes with high data heterogeneity and delayed ground-truth feedback. In
production tasks like recommendation or ad-targeting, the strongest training
signal (the label) often arrives hours after the initial interaction. Existing
systems are \emph{unable} to navigate this: they must either update immediately
using noisy proxy feedback to maintain freshness, or wait for full labels and
suffer extreme model staleness.

\begin{figure}[ht!]
    \centering
    \begin{subfigure}[b]{0.98\columnwidth}
        \centering
        \includegraphics[width=\textwidth]{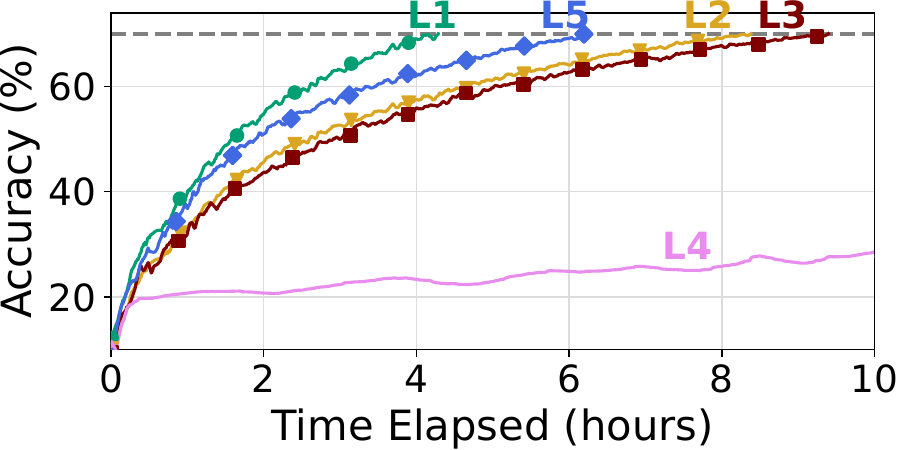}
        \label{fig:background:hetero_unavail_baselines_gap_plot}
    \end{subfigure}
    \caption{Performance gaps in FL widen as we move from ideal conditions (L1) to realistic deployments (L2, L4). Data heterogeneity and client unavailability increase stalls and inefficiency, and methods relying on perfect foresight (L3) break down without it (L4); closing the gap requires real-time, low-cost readiness signals that capture high-value client updates without oracular knowledge (L5).}
    \label{fig:background:hetero_unavail_baselines_gap_plot}
\end{figure}

\subsection{Model Adaptation Gap: Oracular vs. Reality}
\label{sec:background:requirements_gap}

\myparagraph{Opportunity Gap}
To quantify the performance gap with existing FL approaches, we move from an
idealized regime to progressively more realistic deployment settings in
Fig.~\ref{fig:background:hetero_unavail_baselines_gap_plot}, exposing the
technical reasons behind the widening performance gap. While an idealized
baseline of 100\% availability and homogeneous data in Line 1 i.e. L1 ensures
minimal time-to-target accuracy, the introduction of data heterogeneity (L2)
increases gradient variance, increasing the time-to-accuracy even under 100\%
participation. The performance is expected to degrade sharply with realistic
client behaviour, for instance if client availability drops to 50\%. Here rounds
will stall and existing selectors that fail to capture enough informative
updates within tight refresh time windows, will suffer. However, with oracular
or trace-provided client information, despite 50\% unavailability, L3 does not
drop in its performance significanlty compared to L2. The most significant
performance collapse occurs when we withdraw the oracular foresight. Many
state-of-the-art methods like OORT and REFL rely on retrospective traces or
trace-derived availability predictors to estimate future utility  (L4). Without
perfect future knowledge, these systems struggle with low availability,
repeatedly stalling during rounds. Even with over-selection, they fail to
capture the transiently available clients that carry high-value for model
training. Bridging this gap (L5) requires an orchestrator that achieves
near-oracular performance not through foresight, but by exploiting low-cost,
real-time readiness signals to harvest the high-value updates as they appear.

\myparagraph{Design Primitives for Streaming FL}
Closing the empirical gap between Line 4 and Line 5 requires a shift in FL
orchestration infrastructure: availability and freshness must be treated as
\emph{operational control inputs} rather than secondary instrumentation. To
maintain metrics like CTR and conversion in volatile environments, a
streaming-aware FL system must satisfy four design pillars. First, it requires
\textit{low-latency readiness checks} of clients that expose not only
availability but also current data informativeness without saturating network
bandwidth. Second, it must employ \textit{rapid, utility-aware selection} that
fuses a client's recent data utility with its hardware speed capability to
prioritize contributions that most aggressively reduce time-to-target accuracy.
Third, it needs \textit{delay-robust aggregation semantics} to incorporate late,
high-quality updates containing ground-truth feedback, which may arrive hours
after the initial interaction, without allowing staleness to destabilize the
model. Finally, the system must maintain \textit{explicit communication bounds}
to ensure that increased visibility of the client population does not compromise
device battery or server scalability.

Together, these design primitives frame streaming FL as a constrained
optimization problem: \emph{minimize time-to-target accuracy on a live
interaction stream subject to transient availability, evolving utility, frequent
update delays, and bounded communication.} This formulation serves as the
blueprint for the \sys primitives presented in \S\ref{sec:system}.

\section{\sys System}
\label{sec:system}
\subsection{Gaps to handle widespread unavailability}
The discussion above has a recurring pattern: existing FL systems and algorithms
either (A) lack low-cost, near-real-time availability mechanisms, (B) rely on
historic traces or trained predictors that cannot be re-deployed across
applications, or (C) handle staleness with heuristics that aggresively discard
useful updates. This produces three concrete, non-trivial technical problems
that this paper addresses:

\begin{tightitemize}
  \item \textbf{Compact client availability summaries.} How can the server
  obtain a low-cost, privacy-preserving, actionable view of \emph{currently}
  active clients under the application-defined participation criteria? Existing
  tokens/heartbeats are either too coarse (low utilization) or too costly
  (frequent reporting); oracle/trace assumptions are unrealistic in new
  deployments~\cite{bonawitz2019flsysdesign, wang2023flint}.
  \item \textbf{Fresh and evolving client utility estimates.} How can the system
  efficiently refresh inactive clients' utilities so that selectors use
  up-to-date state? Our insight is that utilities can be obtained without full
  training and using lightweight evaluations or probing, but integrating the
  probes into selection while limiting communication is
  challenging~\cite{lai2021oort}.
  \item \textbf{Utility-aware and robust aggregation.} How to combine updates
  with varying staleness at the server so that on-and-off but high-value clients
  are not excluded, while preserving convergence and avoiding undue noise from
  stale updates \cite{nguyen2022fedbuff, abdelmoniem2023refl}.
\end{tightitemize}

\begin{figure}[t!]
  \centering
  \begin{subfigure}[b]{1\columnwidth}
    \centering
    \includegraphics[width=\textwidth]{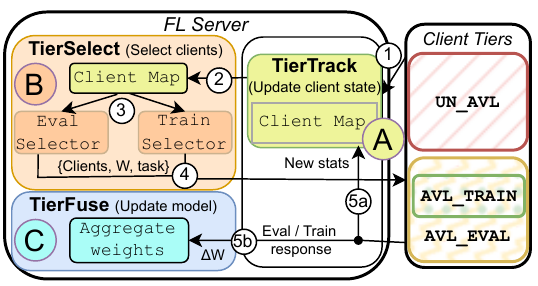}
    \label{fig:system:proposed_system}
  \end{subfigure}
  \caption{\small \textbf{\sys} architecture for FL. \textbf{\sysAware}
    dynamically tiers clients by training capability, \textbf{\sysSample}
    selects and assigns tasks using tier, data utility, and participation
    history, and \textbf{\sysAggregate} weights updates by staleness and
    expected learning contribution.}
  \label{fig:system:proposed_system_overview}
\end{figure}

Motivated by these insigths, we design \sys, a system for robust federated
learning in environments with high client unavailability and heterogeneity. \sys
is the first system to introduce and leverage the concept of \textit{tiered
client pools}. Clients dynamically move among the pools based on their evolving
state (battery, network, data) and capabilities (compute load, speed) over the
long-running FL training jobs. The system can opportunistically select best
suited clients for model training, evaluation of their utility only, or neither,
based on their runtime availability and resource constraints. This approach aims
to maximize each participating client's contribution to the global model with
the goal of minimizing time-to-accuracy and communication costs. \sys achieves
these goals by augmenting the FL paradigm with three mechanisms, illustrated
in~\figref{fig:system:proposed_system_overview}. First is \circled{A}
\textit{\sysAware}, a lightweight, client-driven protocol to indicate each
client's change of tier to the control-plane. \circled{B} Second,
\textit{\sysSample} is a two-path selection mechanism: a TrainSelector that
prioritizes clients to maximize expected learning each round and an EvalSelector
that proactively probes stale or unobserved clients to refresh their utility
estimates and surface occasionally high-value but slower devices. Finally,
\circled{C} \textit{\sysAggregate} judiciously balances each update's quality
with their staleness. The three mechanisms work in cohesion towards the goals,
especially in the presence of high client unavailability and heterogeneity.

\begin{algorithm}[t!]
    \caption{\colorbox{yellow}{\sysAware Algorithm}}
    \label{alg:ClientNotify}
    \begin{algorithmic}[1]

        \Statex \textbf{Input:} $C$: set of all clients \Statex \textbf{Output:}
        real-time client status \State Initialize a data structure $B$ to track
        client availability \While {true} \If{$c_i \in C$ sends a state update}
        \If{state = \colorbox{lightgray}{\texttt{AVL\_TRAIN}}} \State set $B[i]
            \gets 2$ \ElsIf{state = \colorbox{lightgray}{\texttt{AVL\_EVAL}}} \State
        set $B[i] \gets 1$ \ElsIf{state =
            \colorbox{lightgray}{\texttt{UN\_AVL}}} \State set $B[i] \gets 0$
        \If{$c_i$ is currently training} \State checkpoint model for $c_i$
        \EndIf \State cleanup $c_i$ \EndIf
        \EndIf \EndWhile

    \end{algorithmic}
\end{algorithm}

\subsection{Streaming-aware availability tiers}
\label{system:avail_tracking}

A systematic solution is required to bridge the gap between the fluctuating and
low client availability~\cite{mobiperf, wang2023flint, bonawitz2019flsysdesign}
on one hand and the lack of near-real-time availability tracking in existing
systems. As described previously, fresh information about client state is needed
to avoid training stalls and to select the high utility clients. Both combined
help improve the learning progress.
Since the control-plane directly interfaces with the clients, it is most suited
to relay latest client information to the selector.
However, for the mechanism to be effective, it needs to be \textit{real-time},
\textit{accurate}
and \textit{lightweight}, while scaling to thousands of clients.

\par \textbf{\sysAware.} Our approach, described in
Algorithm~\ref{alg:ClientNotify}, segregates clients into one of the three
client pools based on their capability: (i) \cpTrain: available to train and
evaluate, (ii) \cpEval: available only to evaluate, and (iii) \cpUnavl:
unavailable to participate.
While different applications can define these states differently, the clients
would indicate their state to the control plane based on the criteria and their
runtime conditions (compute load, battery status, connectivity, etc). For
example, since training is computationally intensive, a client could be
classified as \cpTrain only when it is connected to a charger and on Wi-Fi (e.g.
OORT, G-FL). However, to tap into under-utilized clients due to this restrictive
criteria, we categorize clients into \cpEval state even when they are not
connected to a charger, but their battery exceeds a certain threshold, since
evaluation is computationally cheaper.

Instead of periodic heartbeats of current state which is communication heavy,
redundant, and unscalable to thousands of clients, the clients proactively
notify the control plane whenever their state changes \circled{1}
(~\figref{fig:system:proposed_system_overview}). To guard against stale or
incorrect state information (e.g., when a client silently drops due to network
or device failure), \sys employs a lightweight check:
Instead of separate messages or a regular heartbeat, it starts a 90-second timer
whenever the aggregator assigns a task; if no response arrives within this
window, the client is marked \cpUnavl.
If the client becomes \cpUnavl during training, it can resume from its last
checkpoint once it transitions back to \cpTrain~\cite{gupta2024jitCheckpt,
eisenman2022checkNRun, mohan2021checkfreq}.

Client state updates are tracked in a simple map data structure that \circled{2}
the control plane forwards to the selector.
\sysAware maintains additional metadata for each client to recognize clients who
rejoin the cohort during the same training round and preclude them from
participating in that round. This is done to prevent the selector from sampling
the same client more than once in a single round, ensuring
fairness~\cite{mcmahan2017communication, nguyen2022fedbuff}.

\begin{figure}[t!]
    \includegraphics[width=0.9\columnwidth]{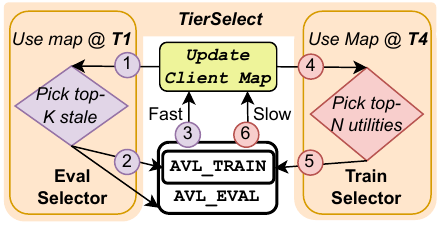}
    \caption{\small \textbf{\sysSample}: lightweight forward-pass evaluations
        refresh stale utilities, the train selector uses these fresh utilities
        to reduce selection error, and the two together accelerate convergence.}
    \label{fig:system:IntelliSample_overview}
\end{figure}

\begin{algorithm}[!ht]
    \caption{\colorbox{yellow}{\sysSample Algorithm}}
    \label{alg:IntelliSelect}
    \begin{algorithmic}[1]

        \Statex \textbf{Global Vars:} $C_T$: clients available for train; $C_E$:
        clients available for eval, but not train; $k_{train}$: train
        concurrency; $K_{train}$: concurrency level; $k_{eval}$: number of
        evaluations; $K_{eval}$: per round evaluation goal

        \Statex \textbf{Outputs:} $S_T$: client selected for train; $S_E$:
        client selected for eval

        \Procedure {\colorbox{pink}{$\mathtt{\sysSample}$}}{\textbf{Global
        Vars}} \While{$k_{train}<K_{train}$} \Comment{async} \State
        $S_T\gets\mathtt{train\_selector}(C_T,k_{train})$ \State $k_{train}\gets
        k_{train}+1$ \State \textbf{return} $S_T$ \EndWhile
        \While{$k_{eval}<K_{eval}$} \Comment{async} \State
        $S_E\gets\mathtt{eval\_selector}(C_E,C_T)$ \State $k_{eval}\gets
        k_{eval}+1$ \State \textbf{return} $S_E$ \EndWhile \EndProcedure

        \Procedure
        {\colorbox{pink}{$\mathtt{train\_selector}$}}{$C_T,k_{train}$}
        \While{true} \Comment{async} \If{recv update from client $c_i\in C_T$}
        \State $k_{train}\gets k_{train}-1$ \State fetch $\mathit{utility}$ of
        $c_i$ \State update $\mathit{utility}$ of $c_i$ in $C_T$ \State sort
        $C_T$ \EndIf \EndWhile

        \While{true} \Comment{async} \State \textbf{return} $C_T[0]$ \EndWhile
        \EndProcedure

        \Procedure {\colorbox{pink}{$\mathtt{eval\_selector}$}}{$C_E,C_T$}
        \While{true} \Comment{async} \State $C_A\gets C_E+C_T$ \If{recv update
        from client $c_i\in C_A$} \State fetch $\mathit{last\_util\_round}$ of
        $c_i$ \State update $\mathit{last\_util\_round}$ of $c_i$ in $C_A$
        \State sort $C_A$ \EndIf \EndWhile

        \While{true} \Comment{async} \State \textbf{return} $C_E[0]$ \EndWhile
        \EndProcedure

    \end{algorithmic}
\end{algorithm}

\subsection{Tiered Client Selection with Fresh Utilities}
\label{system:intelligent_sampling}

Instead of picking random client, recent selectors recognize that \emph{not all
    clients contribute equally} to global progress. A client's contribution
    depends on (i) the richness of its local data (statistical
    utility)~\cite{chen2019federated,peng2019federated,wang2019adaptive,li2020fldirections}
    and (ii) the speed with which it returns updates (system
    utility)~\cite{lai2021oort,huba2022papaya}. A combined \emph{client utility}
    (for example, the product of statistical and system terms) is typically
    estimated after a client participates in training and then used to guide
    future selections.

However, in large, realistic deployments only a small subset of clients is
available at any time and clients are not chosen every round. Consequently,
stored utilities for non-participating clients become stale and no longer
reflect their current potential. Stale utilities induce selection bias.
Frequently high-utility clients are repeatedly favored while other, potentially
useful clients are neglected, slowing convergence and harming fairness across
clients.

Existing approaches offer only partial remedies. OORT inflates utilities for
idle clients to encourage exploration, but this often overestimates inactive
clients' contributions. REFL trains availability predictors and prioritizes
less-available clients, but these models must be retrained for every
application, remain imperfect despite local adaptation, and mis-select clients
during early deployment. As a result, both approaches have limited
applicability.

\sys addresses stale utilities efficiently by proactively collecting fresh
utility measurements from a dedicated, lightweight \textit{eval tier}. Rather
than rely on inference or delayed training replies, \sys obtains up-to-date
statistical utilities via short forward-pass evaluations on \cpEval clients and
feeds these fresh values into selection decisions.

At the core of \sysSample is a compact \textit{ClientMap} that tracks each
client’s latest statistical utility, system metrics, and
\textit{last\_utility\_round}. \sysAware updates pool memberships, while
\sysSample runs two concurrent selectors: a high-frequency
\textit{eval\_selector} and a lower-frequency \textit{train\_selector}
(\figref{fig:system:IntelliSample_overview}). The eval selector sends
lightweight forward-pass evaluations to clients in \cpTrain and \cpEval
prioritized by most stale first; these complete up to $20\times$ faster than
full training~\cite{xu2024fwdllm}, and their returned utilities are written to
the ClientMap. The train selector then uses these refreshed utilities to pick
top-utility trainers at the set concurrency level. Trainers perform full local
training and return model updates and utilities, closing the freshness feedback
loop. Since \textit{last\_utility\_round} is updated after both evals and
training, the ClientMap always holds the freshest client observations.

The eval selector is deliberately rate-limited (assign at most $e$ eval tasks
per round) to avoid stealing training capacity from available clients. This
balances freshness and resource contention: frequent evals refresh utilities
but, if unconstrained, could reduce available trainer choices (during eval task
execution). Each \textit{eval} also adds to the communication cost. However,
timely fresh utilities substantially reduces selection error and more than
amortizes the added eval traffic. For efficiency, selector logic operates only
on the currently available sets (\textit{AVL\_TRAIN}, \textit{AVL\_EVAL})
reported by \sysAware; this keeps rank computations small and ensures only
eligible clients are chosen. The result is a lightweight, fast, and scalable
selection strategy that trades off system and data heterogeneity by leveraging
fresh client utilities amid high client inactivity and heterogeneity.

\begin{figure}[t!]
    \centering
    \begin{subfigure}[t]{0.85\columnwidth}
        \centering
        \includegraphics[width=\textwidth]{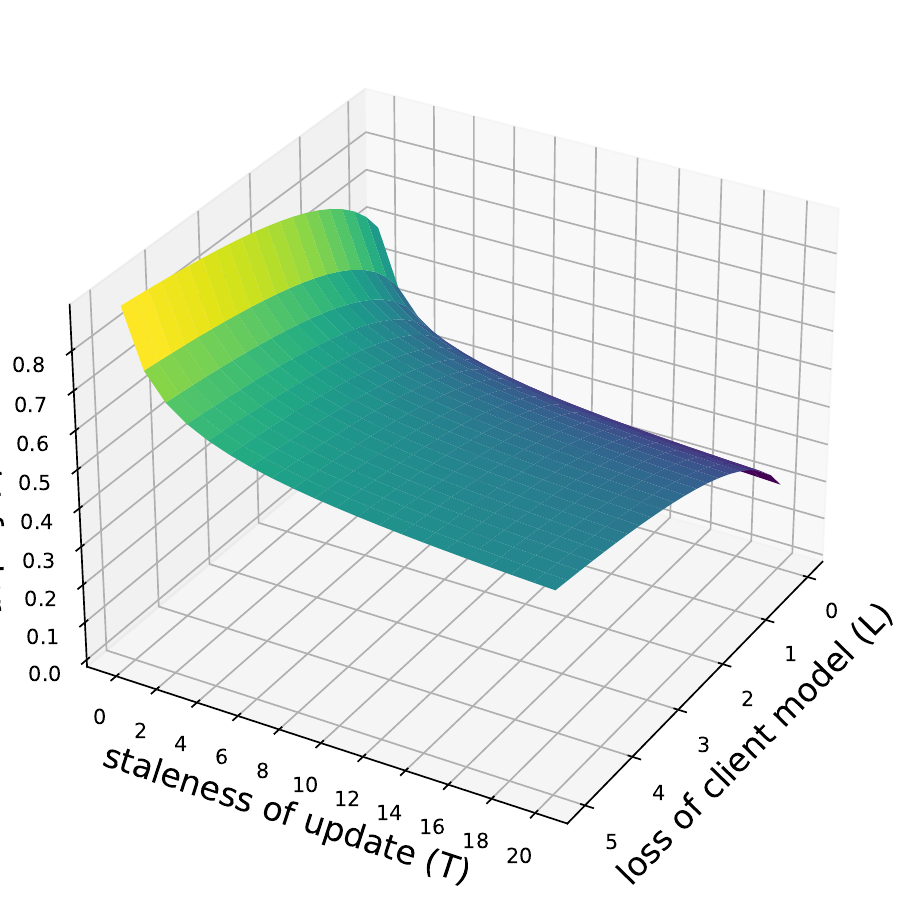}
        \label{fig:system:IntelliAgg_3dplotB}
    \end{subfigure}
    \label{fig:system:IntelliAgg_3dplots}
    \caption{\sysAggregate gives weight based on the staleness and statistical
        utility of the updates, as both are important to gauge the quality of
        the update for the global model.}
\end{figure}

\subsection{Utility-aware, delay-robust aggregation}
\label{system:effective_agg}

Round-based staleness, the gap between the server model version when an update
arrives and the version the client trained on is simple to compute and widely
used in SyncFL and AsyncFL. However, staleness alone conflates two causes of
age: (i) client unavailability and (ii) slow computation or communication.
Neither indicates whether an update is still informative. Prior work often
discounts updates solely by staleness, discarding delayed yet useful
ones~\cite{nguyen2022fedbuff,xie2019asyncFLopt}. \sys distinguishes these cases
by adding a statistical signal: the \emph{utility} of an update, derived from
the client's local loss. A high loss implies the model is poorly aligned with
that client's data and the update remains valuable; a low loss suggests the
update can be safely down-weighted by age.

\sysAggregate encodes the tradeoff between update age and informativeness. Each
update \(u_i\) carries its staleness \(T_i\) and local loss \(L_i\). Two basis
functions: staleness decay \(\alpha(T)\) (decreasing in \(T\)) and utility gain
\(\beta(L)\) (increasing in \(L\)) are combined using a tradeoff parameter
\(\sigma\!\in[0,1]\):
\[
  w_i=\sigma\,\alpha(T_i)+(1-\sigma)\,\beta(L_i),
\]
\[
  \Delta=\frac{\sum_i w_i u_i}{\sum_i w_i}, \qquad
  \theta\!\leftarrow\!\theta+\eta\,\Delta.
\]
Smooth parameterized forms provide interpretable control:
\[
  \alpha(T)=\frac{1}{(1+T)^a}, \qquad
  \beta(L)=\min\!\Big(1.5-\frac{1}{(1+L)^b},1\Big).
\]
Here \(a\) controls how quickly weight decays with age, \(b\) controls how fast
utility saturates, and \(\sigma\) sets the balance between the two. This design
rewards delayed but informative updates (high \(L\)) while still damping very
old or low-utility ones, balancing quality and stability in aggregation.

Two safeguards make \sysAggregate robust in practice. It buffers a short window
of recent updates to stabilize normalization and prevent a single extreme update
from skewing aggregation. Updates older than a threshold \(T_{\max}\) are
clipped or dropped, and optional server-side weight clipping or L2 normalization
further bound any client’s influence. These lightweight measures require no
extra client computation beyond the local loss already produced during training.

Unlike round-based schemes (e.g., REFL’s round-aware approach), \sysAggregate
explicitly combines update age and statistical utility. Staleness penalizes
divergence risk, while high utility rescues delayed but valuable updates. This
yields a simple, interpretable policy suited for heterogeneous, unreliable
deployments—suppressing obsolete noise without discarding late, informative
updates.

\section{Implementation}
\label{sec:impl}
\sys is implemented as a compact set of extensions to Cisco's Flame framework%
\footnote{\url{https://github.com/cisco-open/flame}}, reusing Flame's control
plane, dispatcher, and transport stack. It has three plugins: \sysAware monitors
client activity, capability and maintains lightweight availability pools;
\sysSample implements the two parallel selectors (frequent \textit{eval
  selector} and less-frequent \textit{train selector}) and operates on the small,
available-client view provided by \sysAware; and \sysAggregate is a
buffering-and-weighting layer in the optimizer path that applies the mixed
staleness/utility weights and emits normalized server updates.

State is intentionally minimal and colocated in the control plane: a compact
\textit{ClientMap} stores per-client tuples such as \textit{(stat\_utility,
  system\_metrics, last\_utility\_round, curr\_tier)}. Clients piggyback their
local loss and round duration on replies so selectors and the aggregator can
update ClientMap without extra RPCs. Fast evaluations are implemented as a
single forward pass (empirically upto 20X faster than a full training run) and
write back refreshed statistical utilities to ClientMap, enabling timely,
low-cost updates to selection decisions.

Aggregation is implemented as a configurable-size buffer plus simple
element-wise weighting and normalization. \sysAggregate reads staleness and loss
from the client update, computes weights and produces the server model update;
these operations are cheap. To remain agnostic to communication protocol, we
extended Flame's channel backend with the control semantics needed for
availability and eval requests (the default deployment uses MQTT), and we keep
all extra state in the control plane so that client-side changes are minimal.

Overall, \sys comprises a set of to-be-open-sourced plugins to Flame to deliver
\sysAware, \sysSample, \sysAggregate and ClientMap, all of which cause minimal
runtime disruption and low overhead.

\begin{figure}[t!]
    \centering
    \begin{minipage}[b]{0.75\columnwidth}
        \centering
        \includegraphics[width=\linewidth]{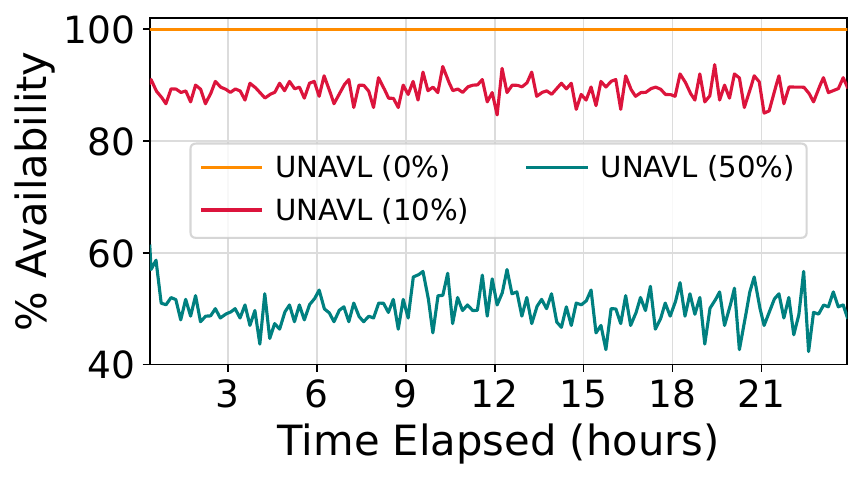}
        \\[2pt]
        \small (a) Synthetic availability pools. \end{minipage}\hfill
    \begin{minipage}[b]{0.75\columnwidth}
        \centering
        \includegraphics[width=\linewidth]{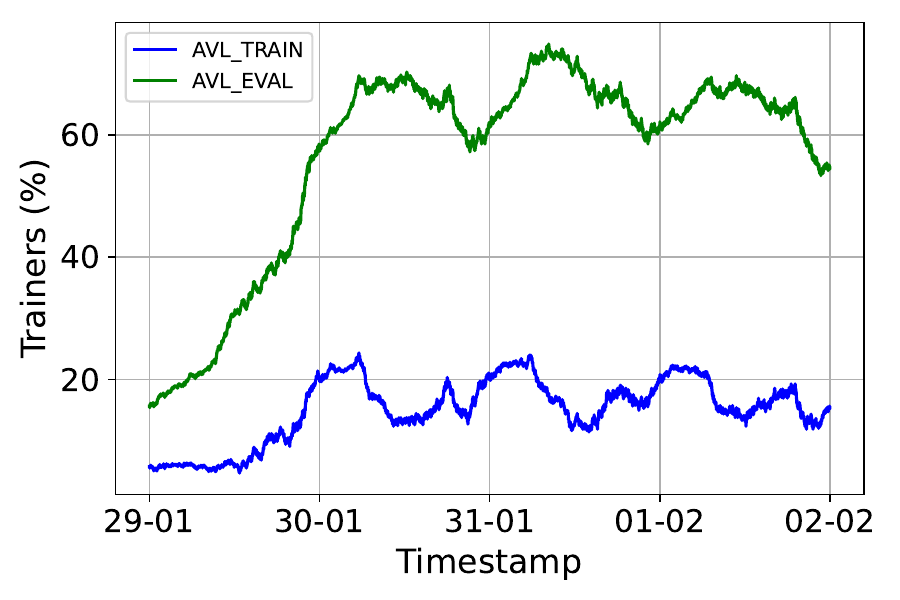}
        \\[2pt]
        \small (b) MobiPerf real-world trace.
    \end{minipage}
    \caption{Client availability traces and three-tier pools. The top panel shows the synthetic client availability distributions used in experiments (high \(\rightarrow\) 100\%, 90\%; medium \(\rightarrow\) 50\%). The bottom panel shows the MobiPerf real-world trace used to stress low-availability regimes. Client pools of $AVL\_TRAIN$ and $AVL\_EVAL$ become especially important here. Traces and the system drives per-client activity of evaluation/training tasks for FL training.}
    \label{fig:traces_synthetic_mobiperf}
    \vspace{-0.1in}
\end{figure}

\begin{table}[t!]
    \centering
    \small
    \setlength{\tabcolsep}{4pt}
    \renewcommand{\arraystretch}{1.01}
    \begin{tabularx}{\columnwidth}{ @{}l
>{\centering\arraybackslash}p{0.12\columnwidth}
>{\centering\arraybackslash}p{0.22\columnwidth}
>{\centering\arraybackslash}p{0.18\columnwidth}
>{\centering\arraybackslash}p{0.18\columnwidth} @{}}
        \toprule
        \textbf{Strategy}          & \textbf{Type}          &
        \textbf{Avail.\,knowledge} & \textbf{Selection}     &
        \textbf{Aggregation}
        \\
        \midrule
        \Oort                      & Sync                   & None & OORT &
                                   FedAvg \\
        {\itshape \OortStar}       & {\itshape Sync}        & {\itshape Oracle }
                                   & {\itshape OORT}        & {\itshape FedAvg }
                                   \\
        {\itshape \OortAsync}      & {\itshape Async}       & {\itshape None} &
        {\itshape OORT}            & {\itshape FedBuff} \\
        {\itshape \OortAsyncStar}  & {\itshape Async}       & {\itshape Oracle}
                                   & {\itshape OORT}        & {\itshape FedBuff}
                                   \\
        \ReflStar                       & Sync                   & Tuned
        Predictor & REFL IPS & REFL SAA \\
        {\bfseries \sys}           & {\bfseries Async}      & {\bfseries
        \sysAware}                & {\bfseries \sysSample} & {\bfseries
        \sysAggregate} \\
        \bottomrule
    \end{tabularx}
    \caption{Overview of strategy configurations used to evaluate \sys against
    key baselines in FL. Regular font indicates prior systems
    (OORT~\cite{lai2021oort}, REFL~\cite{abdelmoniem2023refl}); italic rows are
    adaptations; bold row represents \sys. Oracle ($\ast$) denotes prior
    knowledge of client traces.}
    \label{tab:agg-config}
\end{table}

\section{Experiment Setup}
\label{sec:expt_setup}

\noindent We perform the evaluation
in a cross-device FL testbed built on Flame~\cite{flame2023} with
MQTT~\cite{light2017mosquitto}. The testbed colocates the FL server (control
plane, selector, aggregator) and an emulated population of clients on a
multi-GPU node, following prior emulation-based FL studies (e.g.,
FedScale~\cite{lai2022fedscale}, OORT~\cite{lai2021oort},
REFL~\cite{abdelmoniem2023refl}). Experiments sweep heterogeneity along three
axes: (i) data, (ii) compute / runtime, and (iii) availability, so the results
cover a broad range of realistic operating points. We align Dirichlet
parameters, runtime distributions, and availability scenarios with prior work to
enable meaningful comparisons.

\noindent\textbf{Testbed hardware and runtime emulation.} The host machine with
with 8\,NVIDIA A40 GPUs, 500\,GB RAM, and an AMD EPYC 7513 32-core CPU runs the
FL server and up to 300 emulated clients. Each of the clients are assigned (i)
per-round training delays to emulate compute and network differences, and (ii)
data partitions for data heterogeneity. Per-round client durations are sampled
from a broad distribution spanning \(\approx\)1 to 60\,s, matching the
device-speed and latency distribution reported in production-oriented studies
(Papaya~\cite{huba2022papaya}, FedBuff~\cite{nguyen2022fedbuff}).

\noindent\textbf{Datasets and models.} We evaluate two modalities using two
datasets: \textbf{Image (CIFAR-10)}: 60k images, 10 classes; model is a 12-layer
CNN (\(\approx\)0.5M parameters) and \textbf{Speech (Google Speech Commands
v2)}: \(\approx\)105k one-second clips, 35 labels; model is a ResNet-derived
speech network (\(\approx\)7.2M parameters). All runs start from the same
initialization and use identical configs for parity.

\noindent\textbf{Client populations, data heterogeneity, and availability.} Data
heterogeneity is controlled with Dirichlet partitions \(\mathrm{Dir}(\alpha)\)
for \(\alpha\in\{0.1,1,10,100\}\) (from strongly non-IID to near-IID). To
emulate real-world FL deployments~\cite{ye2023heterogeneous}, experiments below
fix data heterogeneity to \(\alpha=0.1\) unless stated otherwise. It is also the
most challenging regime to stress availability and FL training. The image task
uses 300 clients; the speech task uses 100 clients.

Availability scenarios include synthetic population-level settings (100\%, 90\%,
50\%) and a real-world MobiPerf trace with very low, bursty participation
(\(\sim\)10–22\% available). Figure~\ref{fig:traces_synthetic_mobiperf}
visualizes the synthetic pools and the MobiPerf trace; traces determine
per-client availability across the experiment runtime.

\noindent\textbf{Baselines, selector/aggregator variants, and configuration.} We
compare \sys against existing baseline FL (OORT, REFL), and also against their
variants which incorporate some of \sys's features: \Oort (SyncFL, unaware of
availability), \OortStar (SyncFL + oracle TraceAvail), \OortAsync (OORT extended
to AsyncFL), \OortAsyncStar (OORT AsyncFL + oracle TraceAvail), \ReflStar
(trace-tuned availability predictor based selector + stale aggregations in
SyncFL), and \sys (availability-aware selector + freshness-aware aggregator).
For clarity each strategy is characterized along three axes: availability
knowledge (none / oracle TraceAvail / predictor / \sysAware), selection policy
(OORT / REFL IPS / \sysSample), and aggregation rule (FedAvg SyncFL / REFL SAA /
FedBuff AsyncFL / \sysAggregate), as summarized in Table~\ref{tab:agg-config}.
The oracle TraceAvail is provided only to certain baselines to bound
availability-aware performance; otherwise all strategies share the same
per-client runtime/data settings and local training configurations.

\noindent\textbf{Metrics.} Primary metrics are wall-clock time-to-accuracy
(target: 70\% on both tasks) and total communication (bytes moved). Secondary
analyses include client utility-age distributions, per-round stall durations,
and update-staleness distributions at the aggregator; these are used to
attribute measured performance differences to stalls, utility staleness, or
staleness in updates.

\begin{figure}[t!]
    \centering
    \begin{subfigure}[b]{1.0\columnwidth}
        \centering
        \includegraphics[width=0.98\textwidth]{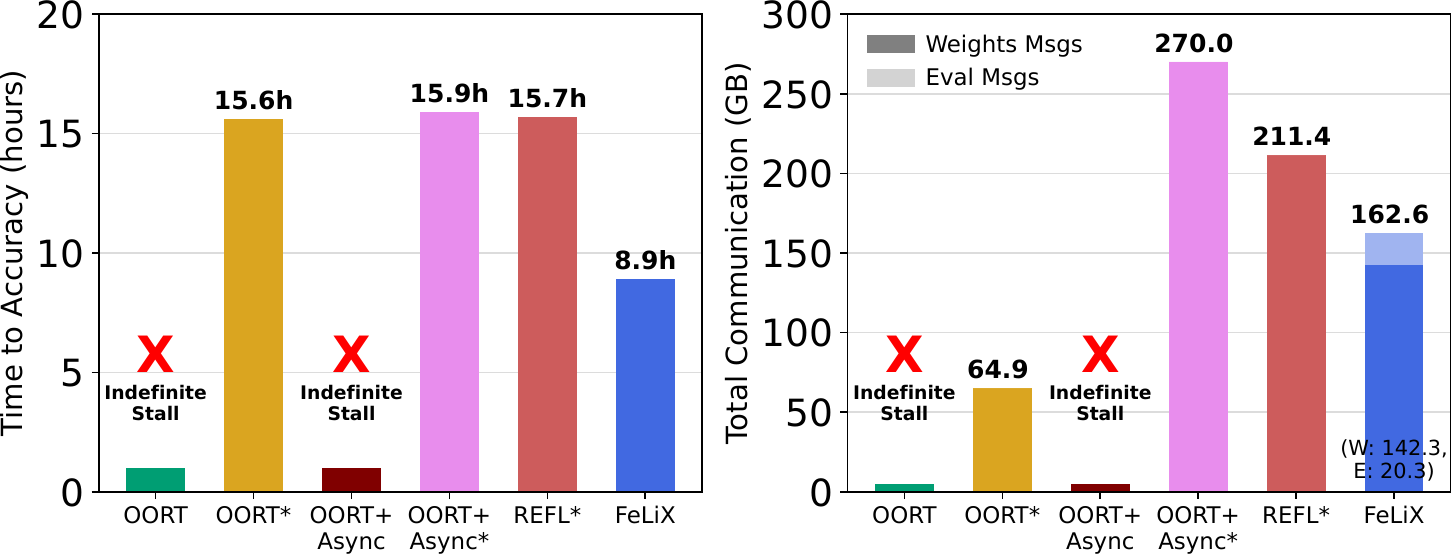}
    \end{subfigure}
    \caption{Time-to-accuracy and communication for CIFAR-10 under the real-world, low availability MobiPerf trace. \sys reaches 70\% accuracy 1.76X faster than oracle-assisted OORT and REFL, while using less total communication (162.6 GB vs 211.4 GB for \ReflStar) by prioritizing more useful client updates early and handling stale updates better.}
    \label{fig:eval:mobiperf_e2e_cifar10}
\end{figure}

\begin{figure}[t!]
    \centering
    \begin{subfigure}[b]{1.0\columnwidth}
        \centering
        \includegraphics[width=0.98\textwidth]{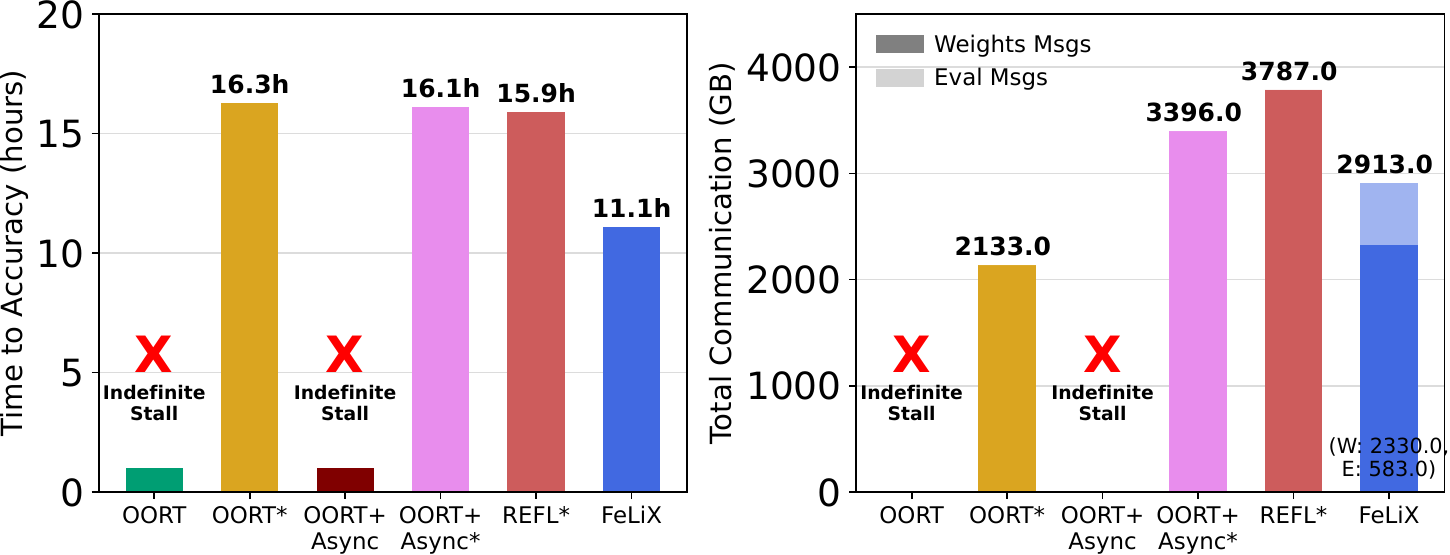}
    \end{subfigure}
    \caption{For Google Speech under the MobiPerf trace, \sys achieves the target accuracy in $\approx$11.1 hours, outperforming oracle-assisted OORT and REFL by $\approx$1.45-1.48$\times$. Despite larger model size and heavier evaluation traffic ($\approx$583 GB), \sys maintains fast convergence with moderate communication cost.}
    \label{fig:eval:mobiperf_e2e_speech}
\end{figure}

\begin{figure*}[ht!]
    \centering
    \includegraphics[width=0.95\textwidth]{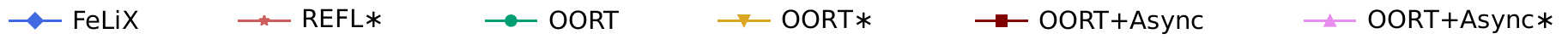}
\end{figure*}

\begin{figure*}[ht!]
    \centering
    \begin{tabular}{>{\centering\arraybackslash}m{0.33\textwidth}
        >{\centering\arraybackslash}m{0.32\textwidth}
        >{\centering\arraybackslash}m{0.32\textwidth}}
        \textbf{100\% Availability}
         & \textbf{90\% Availability} & \textbf{50\% Availability} \\
        \includegraphics[width=\linewidth]{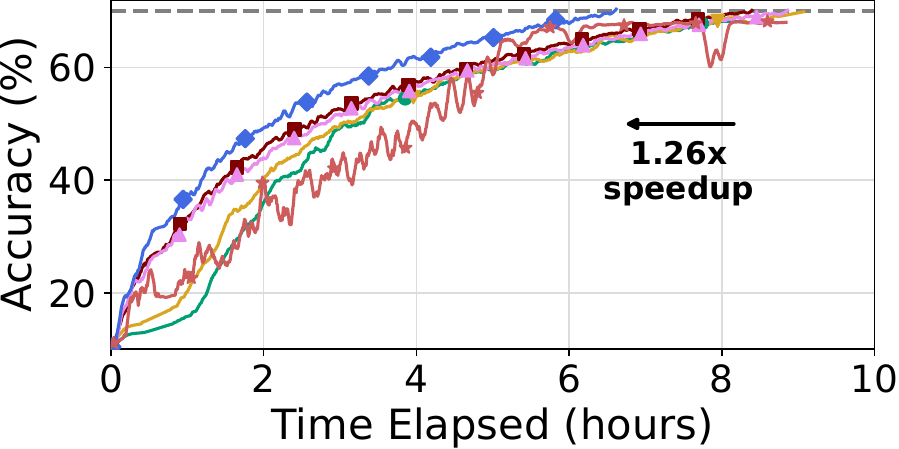}
         &
        \includegraphics[width=\linewidth]{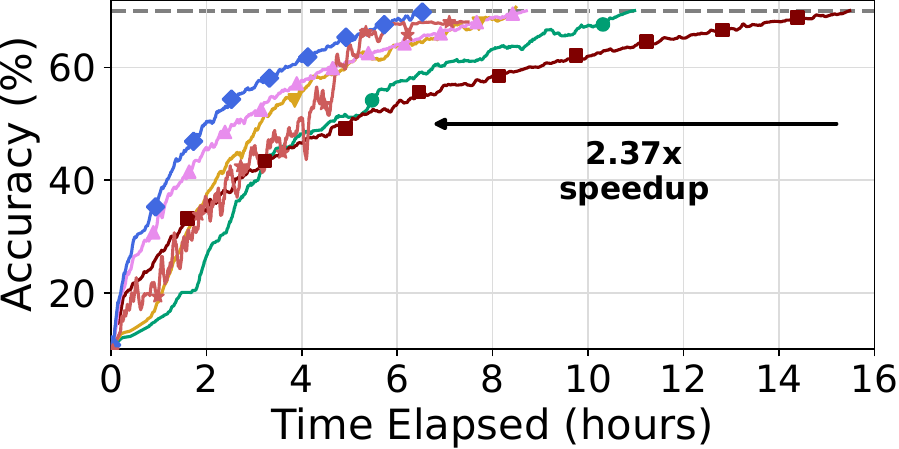}
         &
        \includegraphics[width=\linewidth]{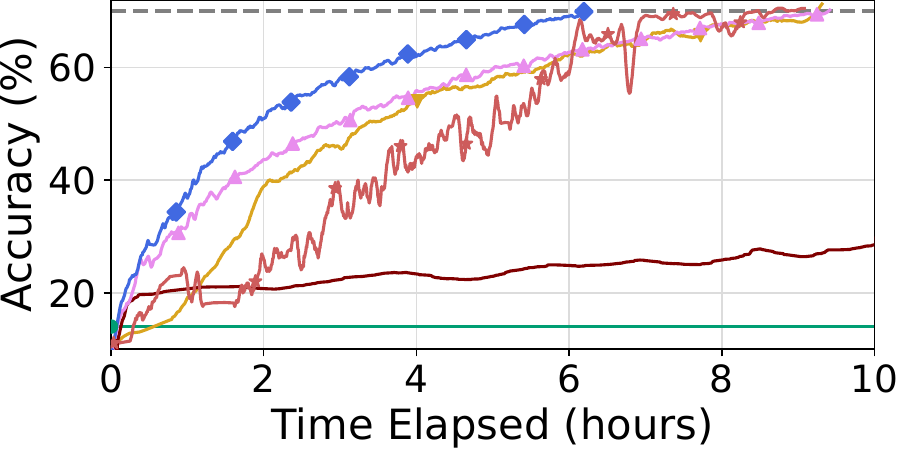}
        \\
        \includegraphics[width=\linewidth]{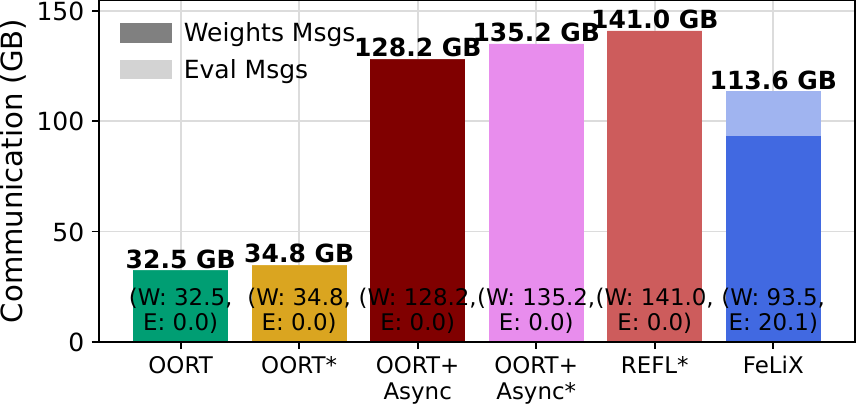}
         &
        \includegraphics[width=\linewidth]{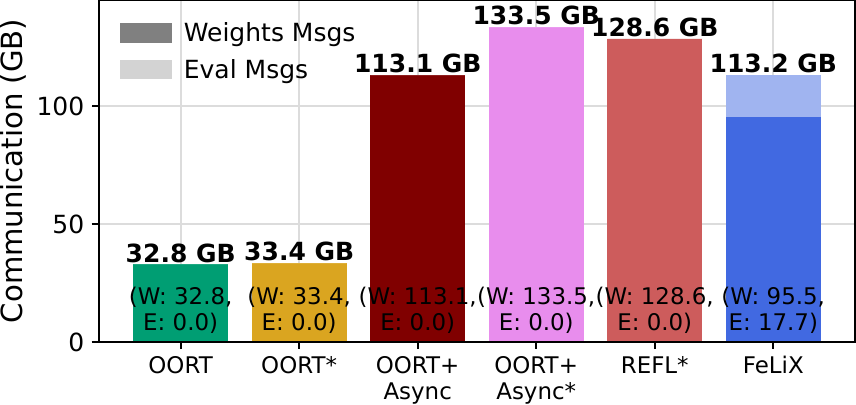}
         &
        \includegraphics[width=\linewidth]{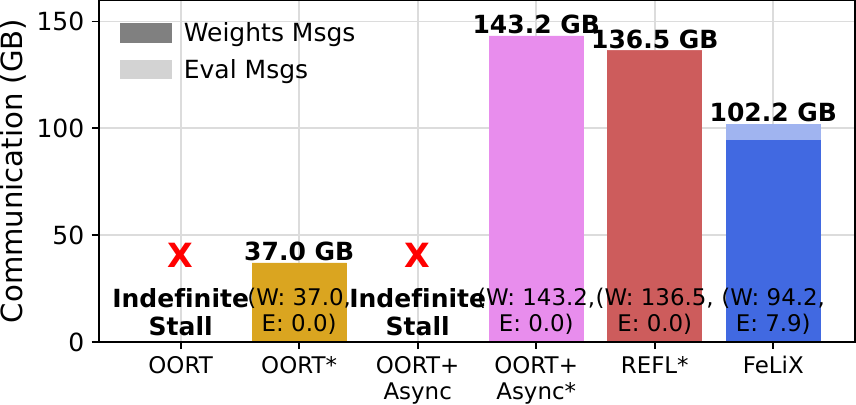}
        \\
    \end{tabular}
    \caption{\sys reaches 70\% accuracy on CIFAR-10 \(>4\times\) faster versus the best non-oracle baseline (\OortAsync) as degree of availability decreases, while also yielding $\approx$40.2\% savings in communication costs. Extent of synthetic availability decreases from 100\% to 90\% to 50\% from left to right. Bottom row captures total communication costs for various strategies.}
    \label{fig:cifar_training_synthetic_comm_grid}
\end{figure*}

\begin{figure}[ht!]
    \centering
    \begin{subfigure}[b]{0.85\columnwidth}
        \centering
        \includegraphics[width=0.95\textwidth]{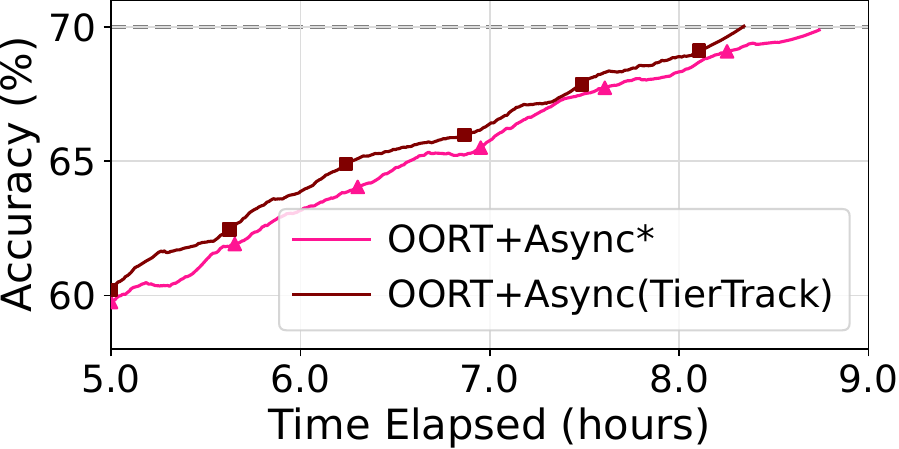}
    \end{subfigure}
    \caption{\small \sysAware provides benefits over oracular knowledge of
        client availability by reducing time-to-accuracy by 5-7\%.}
    \label{fig:cifar_oortAsync_abl_tiertrack}
\end{figure}

\begin{figure}[ht!]
    \centering

    \begin{subfigure}{0.7\linewidth}
        \centering
        \includegraphics[width=\linewidth]{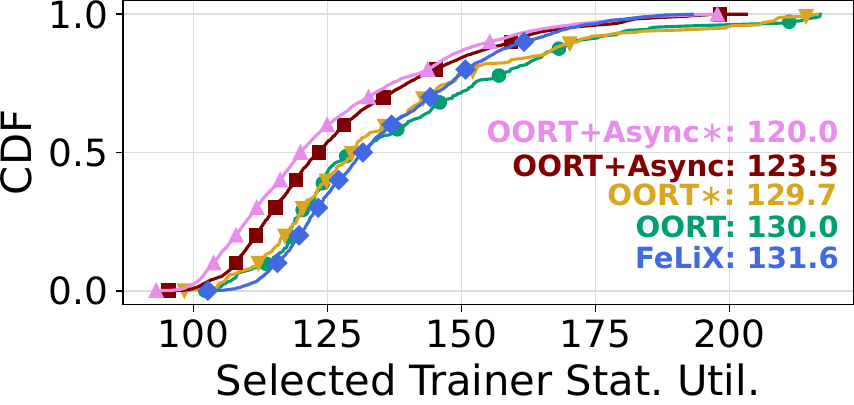}
        \caption{10\% Unavail. Utility (higher is better)}
    \end{subfigure}

    \vspace{0.5em}

    \begin{subfigure}{0.7\linewidth}
        \centering
        \includegraphics[width=\linewidth]{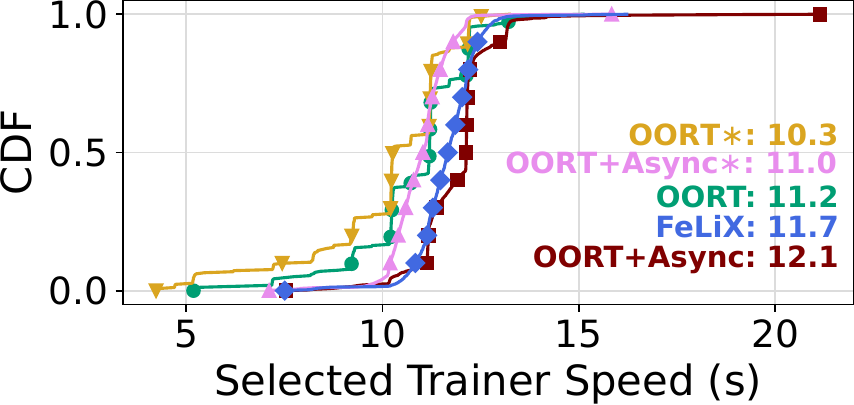}
        \caption{10\% Unavail. Speed (lower is better)}
    \end{subfigure}

    \vspace{0.5em}

    \begin{subfigure}{0.7\linewidth}
        \centering
        \includegraphics[width=\linewidth]{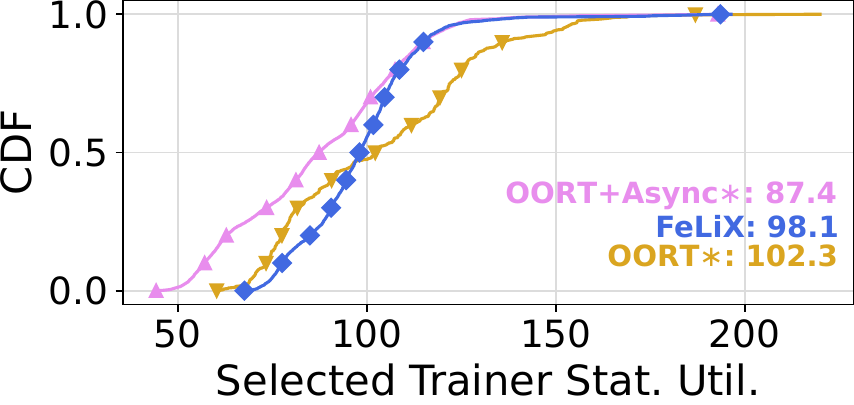}
        \caption{MobiPerf (80–90\% Unavail.) Selector Utility}
    \end{subfigure}

    \vspace{0.5em}

    \begin{subfigure}{0.7\linewidth}
        \centering
        \includegraphics[width=\linewidth]{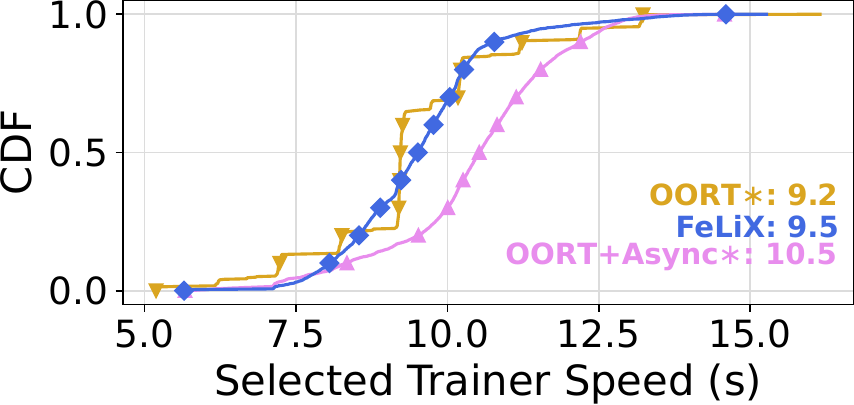}
        \caption{MobiPerf (80–90\% Unavail.) Trainer Speed}
    \end{subfigure}

    \caption{Analysis of selector utility (higher is better) and trainer speed
    (lower is better) across strategies for high and low availability traces.
    \sys maximizes utility while prioritizing fast trainers.}
    \label{fig:cifar_felix_vs_strategies_stat_util_speed_comparison}
\end{figure}

\begin{figure}[ht!]
    \centering

    \begin{subfigure}{0.7\linewidth}
        \centering
        \includegraphics[width=\linewidth]{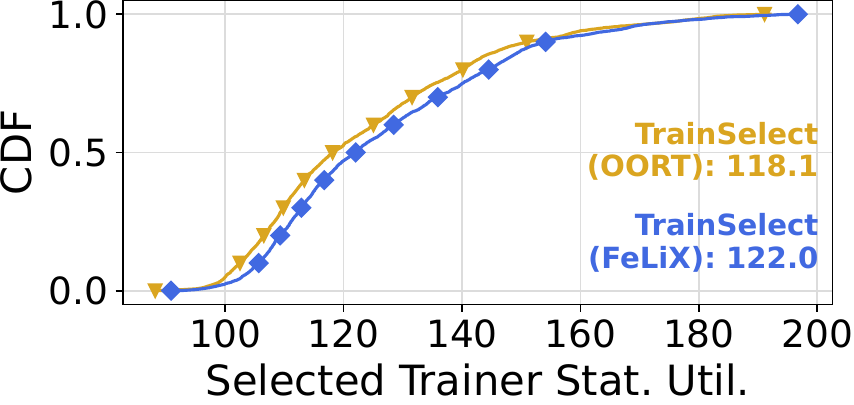}
        \caption{Compare TrainSelectors — Selector Utility (higher is better)}
    \end{subfigure}

    \vspace{0.5em}

    \begin{subfigure}{0.7\linewidth}
        \centering
        \includegraphics[width=\linewidth]{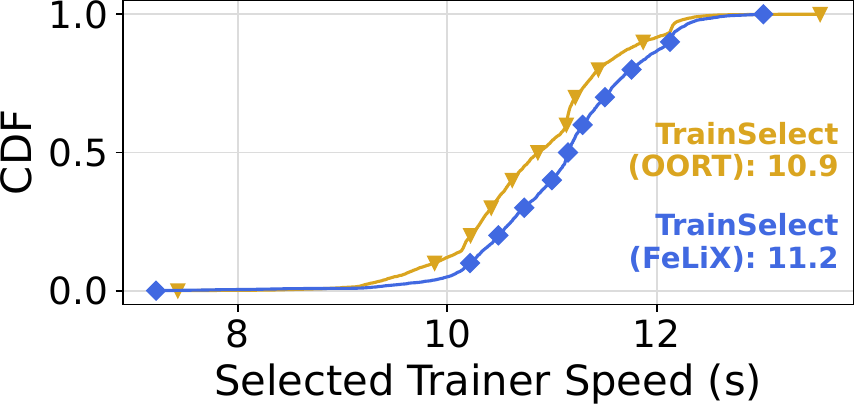}
        \caption{Compare TrainSelectors — Trainer Speed (lower is better)}
    \end{subfigure}

    \vspace{0.5em}

    \begin{subfigure}{0.7\linewidth}
        \centering
        \includegraphics[width=\linewidth]{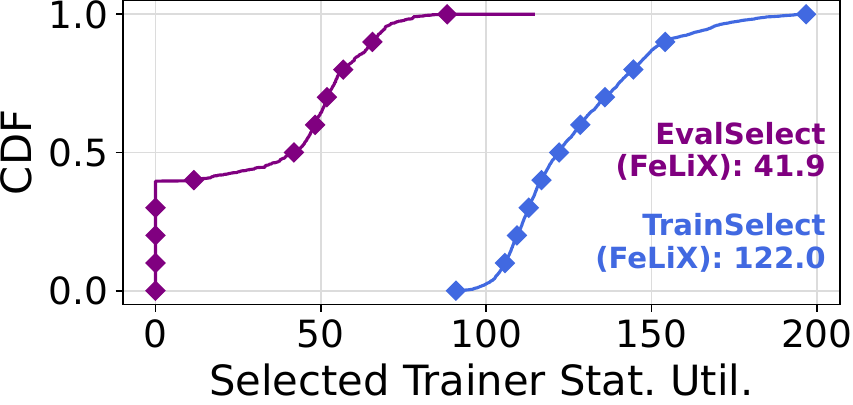}
        \caption{Train vs Eval Selectors — Selector Utility}
    \end{subfigure}

    \vspace{0.5em}

    \begin{subfigure}{0.7\linewidth}
        \centering
        \includegraphics[width=\linewidth]{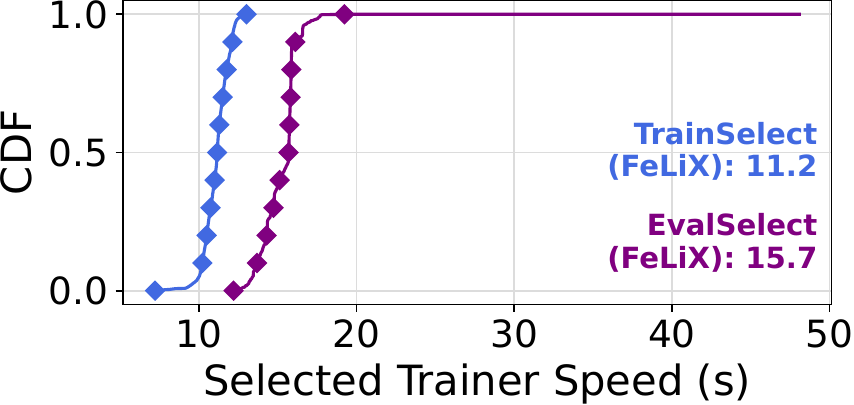}
        \caption{Train vs Eval Selectors — Trainer Speed}
    \end{subfigure}

    \caption{\sys's TrainSelector benefits from the EvalSelector, which probes
    underused clients to reveal high-utility participants, leading to better
    utility-speed tradeoffs.}
    \label{fig:cifar_tierSel_abl_stat_util_speed_comparison}
    \vspace{-0.1in}
\end{figure}

\begin{figure*}[th!]
    \centering
    \setlength{\tabcolsep}{6pt}
    \textbf{\sysAggregate vs FedBuff} \\[0.3em]
    \begin{tabular}{ccc}
        \includegraphics[width=0.32\textwidth]{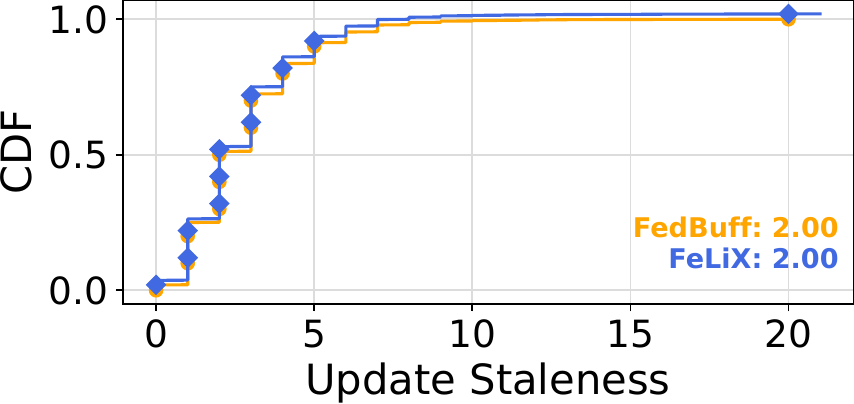}
         &
        \includegraphics[width=0.32\textwidth]{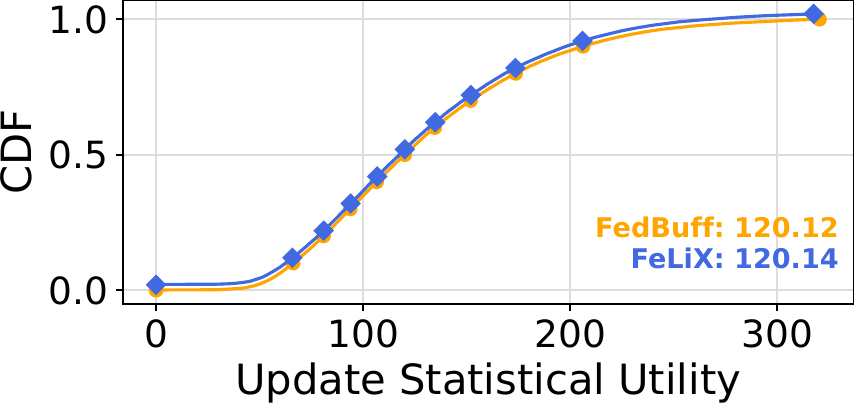}
         &
        \includegraphics[width=0.32\textwidth]{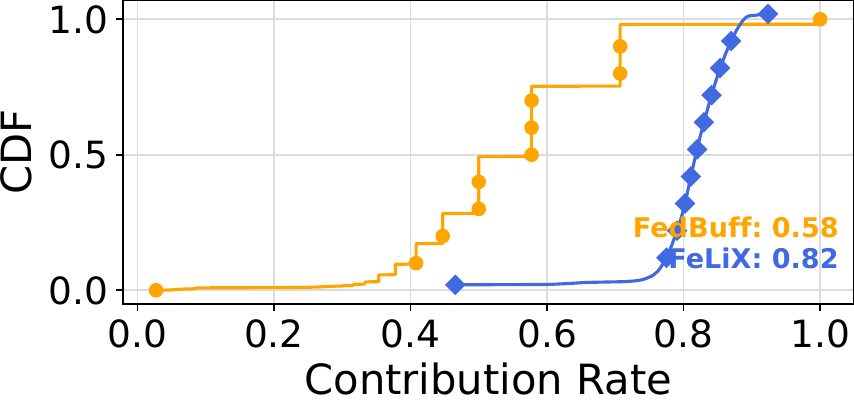}
    \end{tabular}
    \caption{\sysAggregate vs. FedBuff. \sysAggregate prevents the loss of informative signal by maintaining high contribution weights for late updates. By reclaiming gradients from the long-tail of slow or transient clients, \sys reduces time-to-target accuracy compared to the rigid weighting schemes used in current asynchronous baselines.}
    \label{fig:cifar_tierfuse_abl}
    \vspace{-0.1in}
\end{figure*}

\section{Evaluation}
\label{sec:eval}

We seek to answer the following four concrete questions:

\begin{tightitemize}
  \item[\textbf{Q1}] \textbf{End-to-end (real-world):} Can \sys provide tangible
  time-to-accuracy and communication cost benefits under high-unavailability
  real-world traces (MobiPerf), across image and speech modalities?
  \item[\textbf{Q2}] \textbf{End-to-end (synthetic):} Do gains from \sys only
  occur in sparse availability or do they hold across scenarios of moderate
  (50\%) to high (100\%) availability too? This continues to operate in the
  maximal data heterogeneity (\(\mathrm{Dir}(\alpha){=}0.1\)) regime. Can \sys
  reduce wall-clock time-to-accuracy and total communication?
  \item[\textbf{Q3}] \textbf{\sys's components:} What are the end-to-end
  per-component contributions and overheads of \sysAware (real-time
  availability), \sysSample (train + eval-aware selection), and \sysAggregate
  (freshness-aware weighting)? %
  \item[\textbf{Q4}] \textbf{Scalability and robustness:} How do \sys' tracking,
  selection and aggregation %
  perform as clients scale? %
\end{tightitemize}

\subsection{End-to-end: Real-World MobiPerf Trace (Q1)}
\label{eval:e2e_real}

\noindent\textit{Time-to-accuracy.} For CIFAR-10 on the bursty and very low
availability MobiPerf trace, \sys yields an impressive $1.76\times$ speedup by
reaching the 70\% target accuracy versus oracular \OortStar
variants and REFL, the best competing strategies. This reduction in wall-clock
is due to live availability tracking and task-aware sampling. Not only does this
avoid long stalls and reduce blind re-selection, the freshness-aware aggregation
keeps useful updates contributing rather than being ignored. For the speech task
the same pattern holds qualitatively: \sys converges in $\approx$11.1\,h while
oracle-assisted OORT variants and REFL finish in $\approx$15.9–16.3\,h
($\approx$1.45–1.48X slower). The larger absolute times for speech reflect both
model and trace interactions, but the relative improvement of \sys remains
intact across modalities and under the difficult and fluctuating MobiPerf
availability.

\noindent\textit{Communication cost.} The total communication numbers
in~\figref{fig:eval:mobiperf_e2e_cifar10} and
\figref{fig:eval:mobiperf_e2e_speech} must be read together with model sizes and
the mix of evaluation and training task messages. CIFAR-10 uses a small model
($\approx$0.5M parameters) so it requires a modest 1.9\,MB per update sent or
received. On the other hand, the speech model is much larger ($\approx$7.5M
parameters) and needs $\approx$28.6\,MB per exchange. This translates to an
order-of-magnitude larger total bandwidth over the run durations. Under the
MobiPerf trace conditions, all strategies for speech require a large %
(2–3\,TB) of exchange over the course of 3,000 to 6,000 rounds of aggregations
(based on the strategy). In terms of communication cost, %
\sys sits between the \OortAsyncStar and pure sync variants (\OortStar, REFL) in
total bytes while still providing the fastest wall-clock times. For CIFAR-10,
\sys's total communication (162.6\,GB) is still lower than \OortAsyncStar
(270.0\,GB). This gap shows that \sys saves a lot communication cost by
achieving the target accuracy much sooner by selecting and accepting more
helpful client updates early. This includes the extra communication cost of on
\textit{eval} messages ($\approx$20.3\,GB) that keep utilities fresh. For
Speech, \textit{eval} traffic is more consequential: \sys uses $\approx$583\,GB
for evaluation tasks because each evaluation task is now 15X costlier due to the
larger speech model. This %
shows that eval-task bandwidth must be tuned or compressed in large-model
settings.

In short, \sys consistently reduces time-to-target on low availability,
real-world traces for both CIFAR-10 and Speech modality tasks. The
time-to-accuracy and communication-cost is a trade-off: faster convergence often
means more but more useful message exchanges early on, and evaluation overhead
scales with model size. Minimizing end-to-end communication cost therefore
requires balancing evaluation tasks and model-size-aware optimizations (e.g.,
smaller proxy evals or compressed eval uploads) to keep eval overhead small
while preserving the wall-clock gains \sys delivers.

\subsection{End-to-end: Synthetic Traces (Q2)}
\label{eval:e2e_synthetic}

\noindent \textbf{Time-to-Target Accuracy.} Across the synthetic availability
traces in \figref{fig:cifar_training_synthetic_comm_grid}, \sys reaches the 70\%
CIFAR-10 target faster than all baselines, including those with oracular
foresight. Relative to the strongest non-oracle strategy (\OortAsync), \sys
improves time-to-accuracy by $1.26\times$ at 100\% availability, $2.37\times$ at
90\%, and by $>4\times$ at 50\% availability. As availability decreases, the
lines for availability-unaware strategies shift toward the bottom-right, with
time-to-accuracy increasing from 8\,h to over 24\,h. This shift reflects the
heavy cost of incorrect selections and synchronization stalls. While oracular
strategies (\OortAsyncStar) remain stable near 8\,h by avoiding stalls via
trace-knowledge, \sys actually outperforms these oracles. This performance gain
is driven by three synergistic primitives: \sysAware tiering to avoid stalls in
real-time, \sysSample's dual train-and-eval tasks that prioritize utility even
in high-availability regimes, and \sysAggregate, which extracts informative
signal from late updates that other systems would simply discard.

\noindent \textbf{Communication Efficiency and the Cost of Staleness.} \sys
significantly reduces the total communication required to reach target accuracy.
When compared to the oracular \OortAsyncStar, \sys saves approximately $40.2\%$
of total bandwidth. Notably, as unavailability increases, \sys's total
communication cost actually falls by $\approx 9.7\%$, as \sysSample dispatches
fewer evaluation tasks when client utility remains fresher for longer. 

The comparison with \ReflStar reveals a significant efficiency gap. Although
\ReflStar is a SyncFL variant, it maintains a large set of trainers "in-flight"
to accommodate delayed updates and maximize throughput. However, its rigid
policy of discarding any update staler than 5 rounds creates a significant
communication overhead. Empirically, we observe that effectively 25\% of REFL’s
trainer-to-aggregator uplink traffic is discarded for exceeding this threshold.
This is not merely a waste of device uplink traffic; it also represents a loss
of the aggregator-to-trainer downlink bandwidth used to transmit the model
weights initially. Furthermore, this aggressive pruning harms the model itself,
as informative updates from the long-tail of slower or transiently unavailable
clients are lost. By contrast, \sys employs freshness-aware aggregation to
incorporate these updates safely, ensuring that all network communication cost
incurred by the system contributes to the final convergence. This allows \sys to
beat REFL not only in wall-clock time but also in total communication cost
across all availability regimes.

\subsection{\sys Component Benefits (Q3)}
\label{eval:sys_component_contributions}
To decompose the end-to-end gains reported in \S\ref{eval:e2e_synthetic} and
\S\ref{eval:e2e_real}, we inspect three diagnostic metrics at runtime:
aggregator stall duration, client utility age (the latency between a client’s
last evaluation and its current selection), and update staleness at the point of
aggregation. These traces expose the specific failure modes that affect existing
strategies. SyncFL variants without live telemetry suffer from long-tail round
durations as they wait on dropped or straggling clients. Selection algorithms
that do not proactively refresh utility estimates suffer from utility drift,
repeatedly re-selecting historically high-value clients that are currently
unavailable or whose data distribution has shifted. Finally, standard AsyncFL
rules either discard valuable-but-late updates or apply a uniform staleness
penalty, both of which erode the statistical signal. We now show how each \sys
component resolves these failures through a tightly coupled orchestration loop.

\noindent \textbf{\sysAware: Mitigating the impact of client inactivity.}
Integrating \sysAware into the orchestration loop sharply reduces tail latencies
by enabling immediate resampling when selected clients become unreachable. In
our CIFAR-10 experiments, augmenting a stalled \OortAsync baseline with
\sysAware cuts the median (P50) round duration from 258.9s to just 4.9s, while
the P95 tail latency drops from 653.3s to 21s
(\figref{fig:cifar_oortAsync_abl_tiertrack}). This orders-of-magnitude
improvement occurs because \sysAware operates on live telemetry rather than
static traces; it can detect mid-round drop-offs—which affected 6–8\% of all
selections in our traces—and trigger an immediate replacement. Consequently,
\OortAsync+\sysAware achieves a 5–7\% lower time-to-accuracy than even the
trace-informed oracle (\OortAsyncStar), which can avoid stalls but lacks the
mechanism to reactively resample when a selected client disappears
mid-computation.

\noindent \textbf{\sysSample: Maintaining utility freshness via dual-tier
selection.} \sysSample decouples selection into a TrainSelector, which targets
high-progress, responsive clients, and an EvalSelector, which acts as a "scout"
by issuing lightweight evaluation tasks to clients with stale or missing utility
estimates. This dual-tier approach ensures that the server's view of the
population remains accurate despite bursty availability. As shown in
\figref{fig:cifar_felix_vs_strategies_stat_util_speed_comparison}, the
TrainSelector consistently prioritizes clients with higher statistical utility
(median score of 122.0 vs 118.1 for \Oort), even when those clients are
marginally slower (median compute of 11.2s vs 10.9s). This trade-off is
mathematically advantageous: by selecting more informative gradients, the model
requires fewer total rounds to reach the target accuracy. Simultaneously, the
EvalSelector probes low-utility trainers and reveals their true hardware
profiles (P99 $<$ 100), identifying that these clients are significantly slower
(median 15.7s). This fresh evidence prevents the system from making blind,
repeated re-selections of low-yield or hardware-constrained devices.

\noindent
\textbf{\sysAggregate: Regaining information from delayed updates.}
Rather than applying a destructive staleness penalty or discarding late updates,
\sysAggregate computes a contribution factor that balances a client's specific
statistical utility against its update age. This allows the system to
incorporate informative updates from the "long-tail" of the edge ecosystem that
would otherwise be lost. As illustrated in \figref{fig:cifar_tierfuse_abl},
\sysAggregate assigns high weights ($\ge 0.8$) to over 90\% of received updates,
preserving their informative signal even when they arrive moderately late. In
contrast, a FedBuff-style policy, which lacks this informativeness-aware
weighting, assigns such high weights to fewer than 10\% of updates, often
falling back to a flat moderate weight of $\approx$0.5. By effectively
protecting high-value updates from aggressive staleness discounting,
\sysAggregate increases the per-update usefulness of every byte transmitted,
accelerating convergence under the erratic conditions of streaming FL.

\noindent \textbf{\sys: Combined synthesis of components.} These three
components are fundamentally intertwined: \sysAware eliminates stalls to ensure
that EvalSelector's probes reflect actual availability; EvalSelector maintains
fresh utility estimates so TrainSelector can accurately prioritize informative
clients; and \sysAggregate ensures that once an update is received—whether fresh
or moderately stale—its weight reflects its true contribution to the global
model. Together, they transform availability and freshness from passive
observations into active control signals, explaining the substantial wall-clock
and communication gains observed in production-scale settings.

\subsection{Scalability and robustness (Q4)}
\label{eval:sys_microbenchmarks}

\noindent \textbf{Selector and decision latency.} \sys is designed to minimize
server-side overhead by offloading availability tracking to the edge.
Per-decision compute latency remains consistently low ($<0.5s$) because the
server does not perform an exhaustive, synchronous scan of the entire
population. Instead, the client-driven tiering ensures the server only ranks the
active, \emph{ready} subset of the population, while the \textit{EvalSelector}
asynchronously estimates live client utility in the background. This decoupling
allows the orchestration logic to scale gracefully; as the population increases
from 10 to 300 concurrent clients, we observe no significant increase in
selection latency or congestion at the MQTT broker.

\noindent \textbf{Robustness to heterogeneity.} We evaluate the resilience of
\sys across varying degrees of statistical and system heterogeneity by sweeping
Dirichlet distribution parameters ($\alpha \in \{0.1, 1, 10, 100\}$) and runtime
compute distributions (1 to 60 seconds). As expected, the relative gains of \sys
are most pronounced in the strongly non-IID regimes ($\alpha=0.1$) and
low-availability settings. These represent the most challenging and
deployment-relevant scenarios where standard FL systems typically collapse. In
highly heterogeneous environments, the \emph{opportunity cost} of a poor
selection is high; missing a transiently available client with high-utility data
results in a significant loss of gradient progress. \sys's ability to fuse live
availability with freshness-aware aggregation allows it to navigate these sparse
availability environments effectively, ensuring that every successful update
provides maximum statistical gain. This robustness demonstrates that \sys is not
merely faster in idealized settings, but fundamentally more resilient to the
adverse conditions of production edge ecosystems.

\section{Related Work}
\label{sec:related_work}

\myparagraph{FL Applications}
FL was introduced by McMahan \etal~\cite{mcmahan2017communication} as FedAvg,
and~\cite{kairouz2021advances,yang2019federated} survey its landscape and open
challenges. Beyond the production recommender and analytics workloads this paper
targets, FL has been applied to mobile keyboard
prediction~\cite{hard2018federated,chen2019federated}, on-device language
modeling~\cite{xu2023federated,xu2024fwdllm}, healthcare
analytics~\cite{brisimi2018federated,li2019privacy,xu2021federated,nguyen2022federated},
and distribution-shifted domains~\cite{peng2019federated}. Privacy threats
including membership inference~\cite{shokri2017membership} and client-level
differential privacy~\cite{geyer2017differentially} motivate the data-local
design that \sys inherits; its orchestration layer is orthogonal to these
protections.

\vspace{0.5cm}

\myparagraph{FL Convergence, Heterogeneity, and Optimization}
FedAvg convergence under non-IID data~\cite{li2019convergence,wang2020tackling}
identifies data heterogeneity and partial participation as primary theoretical
obstacles; heterogeneous FL settings are surveyed in~\cite{ye2023heterogeneous}.
Adaptive server-side optimizers~\cite{reddi2020adaptive}, matched
averaging~\cite{wang2020federated}, ensemble
distillation~\cite{lin2020ensemble}, and Bayesian non-parametric
fusion~\cite{yurochkin2019bayesian} improve convergence under heterogeneous
data; one-shot FL~\cite{li2020practical} eliminates iterative rounds entirely.
Communication efficiency is targeted by gradient
compression~\cite{konevcny2016federated}, distributed mean
estimation~\cite{suresh2017distributed}, and adaptive communication
scheduling~\cite{wang2019adaptive}. \sys's weighted aggregation policy is
complementary to all of these: it targets update informativeness and model-age
correction rather than modifying the fusion or compression scheme.

\vspace{0.5cm}

\myparagraph{Client Availability, Selection, and AsyncFL}
Section~\ref{sec:background_motivation} analyzes in detail why the leading
utility-aware selectors OORT~\cite{lai2021oort} and REFL~\cite{abdelmoniem2023refl},
buffered AsyncFL~\cite{nguyen2022fedbuff,zhang2023timelyfl}, and production FL
systems~\cite{bonawitz2019flsysdesign,srinivas2025papaya,wang2023flint} fall
short under real-time availability churn. More broadly, async FL has been
studied from decentralized SGD and momentum
perspectives~\cite{lian2018asynchronous,xie2019asyncFLopt,mitliagkas2016asynchrony},
and extended to continual and non-IID
settings~\cite{shenaj2023asynchronous,chen2020asynchronous}.
Staleness handling has received dedicated
attention~\cite{zhou2022towards,rodio2024fedstale}; variance from partial
participation is targeted by FedVARP~\cite{jhunjhunwala2022fedvarp} and
large-cohort analysis~\cite{charles2021large}. Nishio and
Yonetani~\cite{nishio2019client} select clients by deadline compliance under
resource heterogeneity. Prior work~\cite{garg2025client} empirically shows
that ignoring real-time availability substantially increases time-to-accuracy,
directly motivating the \sysAware tracking design in \sys.

\vspace{0.5cm}

\myparagraph{Client Interruption and Local Checkpointing}
A natural response to mid-training client interruptions is to checkpoint local
model state, enabling resumption after conditions recover---a technique
well-developed for distributed DNN
training~\cite{mohan2021checkfreq,eisenman2022checkNRun,gupta2024jitCheckpt,wang2023geminiRecovery}.
In the FL context, however, local checkpointing only defers the problem: the
client still cannot return any update until it becomes available again, and the
global model continues to advance in the interim, rendering saved gradients stale
or model-incompatible on return. \sys addresses the root cause instead, by
exploiting lightweight real-time readiness signals to harvest updates \emph{before}
interruptions occur and by weighting late-arriving updates by their informative
value rather than discarding or blindly accumulating them.

\vspace{0.5cm}

\myparagraph{FL Systems, Frameworks, and Benchmarks}
Production FL platforms~\cite{bonawitz2019flsysdesign,huba2022papaya,srinivas2025papaya}
address large-scale cross-device deployments; FLINT~\cite{wang2023flint}
provides on-device measurement; FedScale~\cite{lai2022fedscale} benchmarks
system performance with realistic traces. Research frameworks
Flower~\cite{beutel2020flower} and FedML~\cite{he2020fedml} support controlled
experimentation. \sys is built on the Flame orchestration
layer~\cite{flame2023} and evaluated against real MobiPerf
traces~\cite{mobiperf}; unlike offline benchmark or framework tools, it targets
production-scale streaming workloads with bursty, minute-level availability
dynamics.

\section{Conclusion}
\label{sec:conclusion}
As production Federated Learning moves from multi-day batch refreshes toward
hourly streaming adaptation, existing orchestration models have proven
insufficient. The interlocking challenges of transient client availability,
shifting data utility, and delayed ground-truth feedback create a freshness gap
that degrades critical user-facing metrics. This work introduced \textbf{\sys},
a redesign of FL orchestration that treats availability and utility not as
passive observations, but as high-frequency control inputs. By unifying three
synergistic primitives: \textit{streaming-aware availability tiers},
\textit{fresh-utility selection}, and \textit{informativeness-aware,
delay-robust aggregation}, \sys bridges the gap between idealized oracular
theory and the realities of the practical FL deployments in the mobile-edge
ecosystem. Our evaluation across multiple modalities and realistic MobiPerf
traces demonstrates that \sys achieves near-oracular performance, delivering a
$2.37\times$ reduction in time-to-target accuracy while simultaneously saving
$1.30\times$ in total communication bandwidth compared to state-of-the-art
baselines. These results confirm that by eliminating synchronization stalls and
reclaiming signal from the long-tail of delayed updates, \sys ensures that
deployed models remain in sync with volatile live data distributions.
Ultimately, \sys provides the necessary framework for high-scale,
privacy-preserving applications to adapt at the speed of user interaction.

\bibliographystyle{acm}
\bibliography{references}

\end{document}